\pdfoutput=1

\documentclass[conference]{IEEEtran}
\IEEEoverridecommandlockouts
\usepackage{cite}
\usepackage{amsmath,amssymb,amsfonts}
\usepackage{algorithmic}
\usepackage{graphicx}
\usepackage{textcomp}
\usepackage{xcolor}
\def\BibTeX{{\rm B\kern-.05em{\sc i\kern-.025em b}\kern-.08em
    T\kern-.1667em\lower.7ex\hbox{E}\kern-.125emX}}

\usepackage{algorithm}
\usepackage{algorithmic}
\usepackage[algo2e,ruled,linesnumbered]{algorithm2e}

\usepackage{booktabs}
\usepackage{multirow}
\usepackage{enumitem}
\usepackage{tabularx}
\newcolumntype{Y}{>{\small\centering\arraybackslash}X}
\usepackage{url}
\usepackage{subfigure}

\def\method{KBGNN}

\begin{document}

\title{Kernel-based Substructure Exploration \\for Next POI Recommendation
}



\author{
\IEEEauthorblockN{Wei Ju$^{1*}$, Yifang Qin$^{2*}$, Ziyue Qiao$^3$, Xiao Luo$^{4,\dagger}$, Yifan Wang$^1$, Yanjie Fu$^5$, Ming Zhang$^{1,\dagger}$}
\IEEEauthorblockA{
$^1$\textit{School of Computer Science, Peking University, China}\\
$^2$\textit{School of EECS, Peking University, China}\\
$^3$\textit{Artificial Intelligence Thrust, The Hong Kong University of Science and Technology (Guangzhou), China}\\
$^4$\textit{School of Mathematical Sciences, Peking University, China}\\
$^5$\textit{Department of Computer Science, University of Central Florida, USA}\\
\{juwei,qinyifang,xiaoluo,yifanwang,mzhang$\_$cs\}@pku.edu.cn,
zyqiao@ust.hk, yanjie.fu@ucf.edu
}
\thanks{$^*$ Equal contribution with an alphabetical order.}
\thanks{$^\dagger$ Corresponding authors.}
}

\maketitle

\begin{abstract}

Point-of-Interest (POI) recommendation, which benefits from the proliferation of GPS-enabled devices and location-based social networks (LBSNs), plays an increasingly important role in recommender systems. It aims to provide users with the convenience to discover their interested places to visit based on previous visits and current status. Most existing methods usually merely leverage recurrent neural networks (RNNs) to explore sequential influences for recommendation. Despite the effectiveness, these methods not only neglect topological geographical influences among POIs, but also fail to model high-order sequential substructures. To tackle the above issues, we propose a Kernel-Based Graph Neural Network (\method{}) for next POI recommendation, which combines the characteristics of both geographical and sequential influences in a collaborative way. \method{} consists of a geographical module and a sequential module. On the one hand, we construct a geographical graph and leverage a message passing neural network to capture the topological geographical influences. On the other hand, we explore high-order sequential substructures in the user-aware sequential graph using a graph kernel neural network to capture user preferences. Finally, a consistency learning framework is introduced to jointly incorporate geographical and sequential information extracted from two separate graphs. In this way, the two modules effectively exchange knowledge to mutually enhance each other. Extensive experiments conducted on two real-world LBSN datasets demonstrate the superior performance of our proposed method over the state-of-the-arts. Our codes are available at https://github.com/Fang6ang/KBGNN.

\end{abstract}


\section{Introduction}

Next Point-of-Interest (POI) recommendation~\cite{lian2014geomf,yao2016poi,liu2014general,li2018next,han2020stgcn,kim2021dynaposgnn} has raised intensive attention in recent years due to the rapid growth of Location-based Social Networks (LBSNs), such as Yelp, Facebook Places, and Foursquare. Such location-based services generate large volumes of historical check-in sequences, which are valuable for service providers to analyze user behavioral patterns in making every decision and thus recommend the next POI a user may want to go. Therefore, next POI recommendation plays a vital role in enhancing user experience in both search efficiency and new interest discovery, and has wide applications including location-based advertising, route planning, and online food delivery.

By analyzing and understanding user mobile behaviors and preferences of massive historical check-in sequences, there are many approaches that have been proposed to make a personalized POI recommendation from various aspects. Traditionally, early models mainly focus on matrix factorization (MF)~\cite{lian2014geomf,yao2016poi} and markov chains (MC)~\cite{rendle2010factorizing,cheng2013you}. To better model the sequential dependencies of user POI sequences, researchers adopt recurrent neural networks (RNNs) and their variants~\cite{feng2018deepmove,li2018next,manotumruksa2018contextual} to better characterize the long periodic and short sequential features of user trajectories. More recently, various models~\cite{han2020stgcn,zhao2020go,fan2021meta} have exploited temporal and spatial relations between movements to alleviate the data sparsity of user check-in data. Besides, many state-of-the-art methods enhance the quality of POI recommendation via incorporating attention mechanism~\cite{luo2021stan,huang2021hgamn} or knowledge graph ~\cite{cui2021sequential} to integrate this auxiliary information.

\begin{figure}[t]
    \centering
    \includegraphics[width=\linewidth]{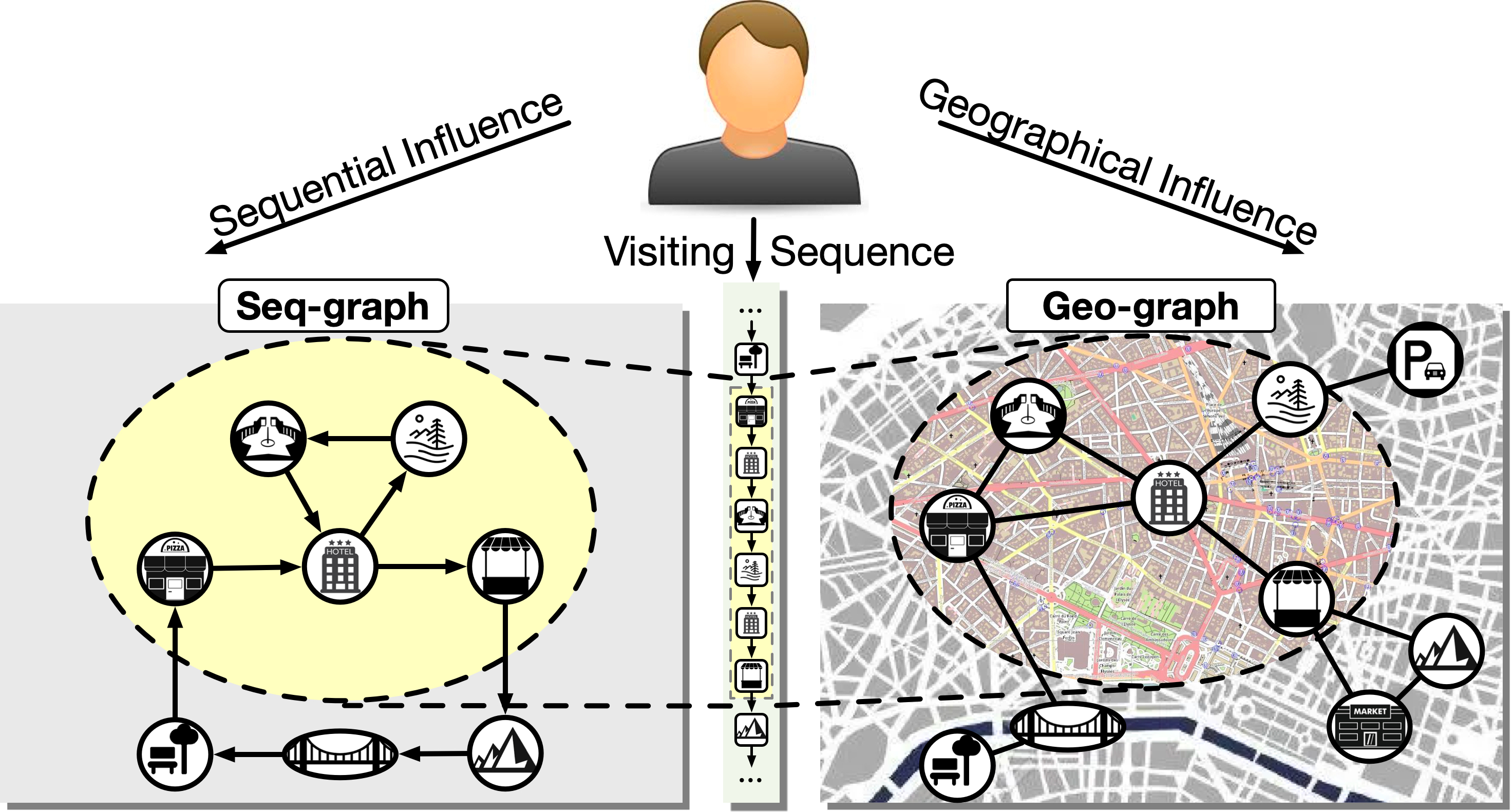}
    \caption{An illustration of geographical and sequential influences behind the check-in behavior of a user. An observed visit to POIs is affected by both the user's historical sequence and the nearby POIs.}
    \label{fig:illustration}
\end{figure}

Although the aforementioned approaches acquire encouraging performance, it is of great significance to explore the major driving forces of next POI recommendation. In fact, we argue that geographical influence and sequential influence are two key factors in POI recommendation. As shown in Fig.~\ref{fig:illustration}, the location ``hotel'' that the user visit is influenced by its neighboring nodes from sequential and geographical graphs. 
From the view of sequential graphs, users tend to re-visit familiar POIs that have appeared in their behavior sequences. For example, frequent visits to theaters may reveal a user's taste in art, which suggests that similar theaters are more likely to be the next POI to visit. From the geographical graph view, POIs with a short physical distance are intuitively more attractive to users. For instance, a user who has just come out of a downtown shopping mall has more chances to visit a fast-food shop nearby for dinner, rather than his/her favorite restaurant in the countryside. In summary, the two influences affect user behaviors and preferences jointly. It naturally raises a meaningful question: \emph{Are existing methods capable of fully capturing these two key factors?}


To answer this question, we need to deepen our understanding of existing methods. Despite the versatility of these POI recommenders, most existing approaches still suffer from two key limitations:
\textbf{(i) Inability to explicitly explore topological geographical influences among POIs}. Most methods usually regard geographical influences as spatial information and incorporate this auxiliary knowledge into RNN-based architectures. In addition, these approaches merely capture geographical influences depending on physical distance or successive relations among POIs. They are incapable of exploring the complex and topological geographical influences among POI networks, which are essential for understanding user preferences. \textbf{(ii) Fail to model high-order sequential substructures}. Most works typically adopt RNN-based methods to model the sequential dependencies of historical user behaviors. However, existing methods may fall short in capturing high-order sequential substructures, which generally reflect personalized user preferences. In our case, the high-order sequential substructures could be several regular routines. For example, a user may only visit three or four spots in succession and then repeat this process, which may result in a `triangle' or a `square' on the map, respectively. There can be also other high-order patterns due to various user preferences. 
As such, we expect an approach which can better explore topological geographical influences among POIs and meanwhile model high-order sequential substructures.

Towards this end, this paper proposes a joint kernel-based graph neural network for next POI recommendation (\method{}), which integrates the characteristics of geographical and sequential influences in a collaborative way. The key idea of \method{} is to combine the advantages of both worlds, and further enhance user experience in both search efficiency and new interest discovery. 
To achieve this goal effectively, we introduce two modules, i.e., a geographical module and a sequential module. For the geographical module, we construct a geographical graph and leverage a message passing neural network to capture the topological geographical influences. For the sequential module, we explore high-order sequential substructures in the user-aware sequential graph using a graph kernel neural network to capture user preferences. It is ideal to combine two modules in a complementary way to enhance recommendations. To effectively train both modules, we introduce a consistency learning framework to jointly incorporate geographical and sequential information extracted from two separate graphs. In this way, the two modules effectively exchange joint knowledge to mutually enhance each other. We validate the effectiveness of \method{} on two real-world datasets. Extensive experimental results demonstrate that our proposed \method{} model achieves high recommendation accuracy and robustness to data sparsity.


The main contributions of this paper are as follows:
\begin{itemize}
\item We propose a joint kernel-based graph neural network for next POI recommendation, which consists of a geographical module and a sequential module to well explore topological geographical influences and meanwhile model high-order sequential substructures.
\item We develop a consistency learning framework to combine the advantages of the geographical module and sequential module jointly, such that they can mutually enhance each other via knowledge communication.
\item We conduct comprehensive experiments to evaluate the performance of the proposed \method{} model over two real-world datasets. The results show the superiority of our \method{} model in POI recommendation by comparing it with the state-of-the-art techniques.
\end{itemize}

\section{Related Work}
\label{sec::related}

\begin{figure*}[!t]
    \centering
    \includegraphics[width=\linewidth]{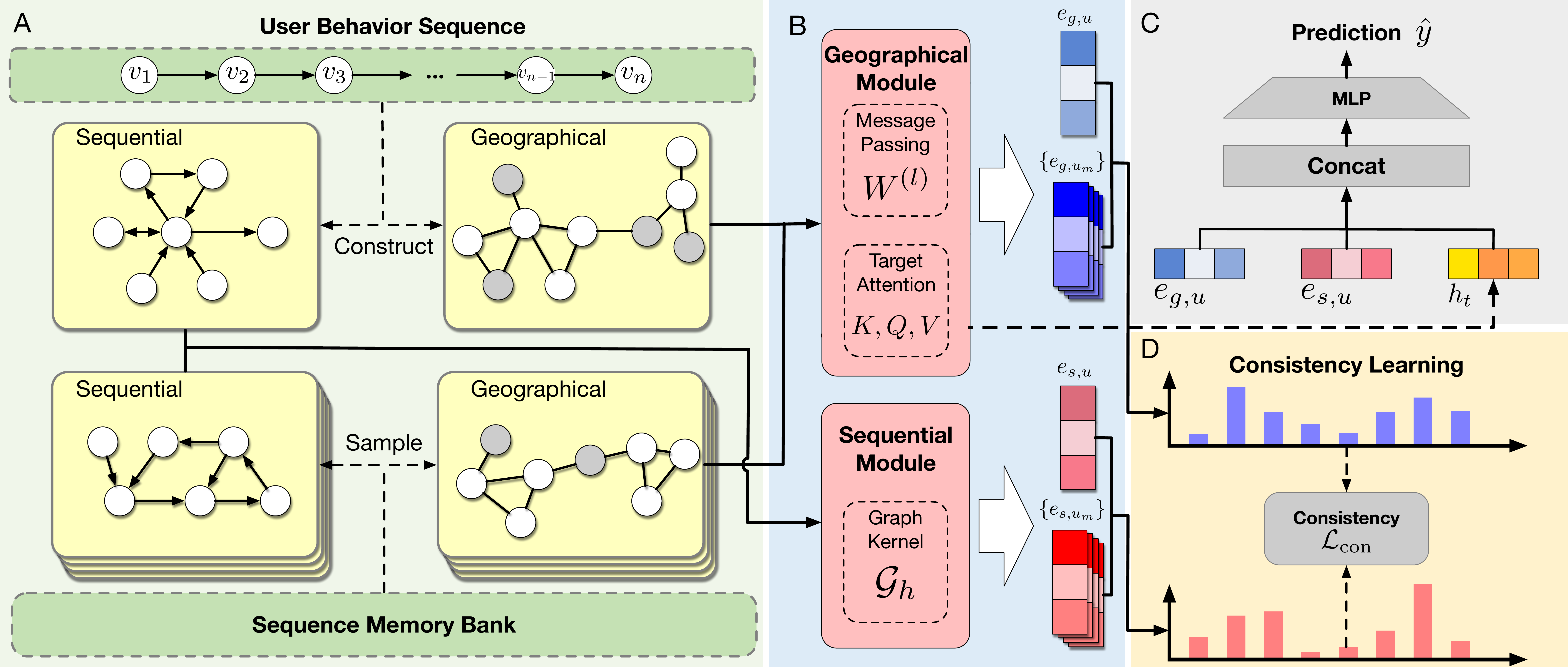}
    \caption{Illustration of the proposed \method{} framework. A) Construction of geographical and sequential graphs; B) Geographical and sequential modules that encode two corresponding graphs and generate the semantic representations; C) Calculate $\mathcal{L}_{\text{con}}$ to encourage the consistency between two modules; D) Prediction layer that combines the two semantic representations and target POI embedding $h_t$ to make CTR predictions.}
    \label{fig:framework}
\end{figure*}


\subsection{Next POI Recommendation}
The next POI recommendation has been an important topic in location-based services, which aims to recommend the next possible visited POI for users based on their historical check-in sequences. Existing approaches for next POI recommendation can be divided into Matrix Factorization (MF) based~\cite{lian2014geomf,yao2016poi,liu2014general} and Neural Network based~\cite{feng2018deepmove,li2018next,manotumruksa2018contextual,han2020stgcn,zhao2020go,fan2021meta}. Most traditional methods are based on MF and the main purpose is to factorize the user-POI interaction matrix to approximately learn user and item latent representation respectively. 
Recent POI recommendation models are mainly based on recurrent neural networks (RNNs) and their variants~\cite{feng2018deepmove,li2018next,manotumruksa2018contextual}, which achieve wide attention due to their superior performance.
DeepMove~\cite{feng2018deepmove} firstly leverages a recurrent layer to learn short-term sequential regularity from highly correlated trajectories, and then captures long-term periodicity with an attention layer.
The state-of-the-art models~\cite{wang2018exploiting,sun2020go} further introduce spatiotemporal information to improve model performance. LSTPM \cite{sun2020go} leverages geo-dilated RNNs to fully explore the temporal and spatial correlations among POIs. Nevertheless, spatial information is typically derived from user behavior sequences and often fails to be explicitly captured from the geographical graph. With \method{}, besides exploring topological geographical influences among POIs from the geographical graph, we also benefit from sequential graph to capture high-order sequential substructures.


\subsection{GNN-based Recommendation}
Graph Neural Networks (GNNs) have become widely acknowledged powerful architectures for modeling recommendation data. GNN-based methods~\cite{kipf2017semi,velivckovic2018graph,luo2022dualgraph,luo2022clear} are founded on the information aggregation mechanism of GNNs. By stacking GNN layers, the model is empowered to capture the complex and high-order graph connectivity. To capture high-order collaborative user-item signals, NGCF~\cite{wang2019neural} and its variant LightGCN~\cite{he2020lightgcn} leverage the high-order relations on the interaction graph to enhance the recommendation quality. DSGL~\cite{chu2021dynamic} uses dynamic sequential graphs to capture the evolutionary dynamics of user's behavior sequences. GEAPR~\cite{li2021you} integrates various factors with an attention mechanism to improve the recommendation interpretability. GSTN~\cite{wang2022graph} incorporates user spatial and temporal dependencies based on graph embedding. Despite effectiveness, our work aims at combining the advantages of geographical and sequential influences in a collaborative way, while their works fail to model two different high-order interaction information jointly via GNNs.

\section{Problem Formulation \& Preliminary}
\label{sec::definition}

\subsection{Problem Definition}


\smallskip
\noindent\textbf{Definition 1: (Point-of-Interest)} A POI is a spatial site (e.g., a restaurant) associated with two attributes: a unique identifier $v$ and geographical coordinates $(longitude, latitude)$ tuple, i.e., $(lon_v, lat_v)$.

\smallskip
\noindent\textbf{Definition 2: (Check-in Sequence)} Denote the user set  $\mathcal{U}=\{u_1,u_2,...,u_{|\mathcal{U}|}\}$, and the POI set $\mathcal{V}=\{v_1,v_2,...,v_{|\mathcal{V}|}\}$. For each user $u\in\mathcal{U}$, his/her historical check-in sequence is organized into a list $s_u=[v_{u,1},v_{u,2},...,v_{u,t-1}]$, containing the $t-1$ POIs he/she has checked-in before, sorted by timestamp.

\smallskip
\noindent\textbf{Problem Statement.} Given a set of POIs $\mathcal{V}$ and a set of users $\mathcal{U}$, each $u\in\mathcal{U}$ has check-in sequence $s_u$. This paper studies location-based click-through-rate (CTR) prediction for next POI recommendation, which aims to leverage geographical and sequential information in user's historical behaviors. Given user $u$ and target POI $v$, the goal of location-based CTR prediction is to predict the probability for $u$ to check-in $v$ next, formulated as $\hat{y}_{uv}=F(u,v|s_u;\theta)$, where $F$ is the learnable function parameterized by $\theta$.

\subsection{Graph Kernels}
\label{sec::gk}

Graph kernels are originally introduced by \cite{gartner2003graph}, which are instances of the R-convolution framework~\cite{haussler1999convolution}, and the basic idea is to decompose graphs into high-order substructures and capture the graph similarity via kernel functions.

\smallskip
\noindent\textbf{Definition 3: (Graph Kernels)}
Given two graphs $G=(\mathcal{V},\mathcal{E})$ and $G'=(\mathcal{V'},\mathcal{E'})$, the graph kernel $K(G, G')$ measures the similarity between them and is defined as:
\begin{equation}
    K\left(G, G'\right)=\sum_{v \in \mathcal{V}} \sum_{v' \in \mathcal{V'}} k_{base}\left(f_{G}\left(v\right), f_{G'}\left(v'\right)\right),
\end{equation}
where the base kernel $k_{base}$ is used to compare substructures centered at nodes~$v$ and $v'$ (i.e., the inner product on Hilbert space), and $f_G(v)$ is the feature vector counting the number of appearances of each substructure (e.g., subtrees~\cite{shervashidze2011weisfeiler}, random walk~\cite{kashima2003marginalized,ju2022tgnn}, paths~\cite{borgwardt2005shortest}, graphlets~\cite{shervashidze2009efficient}) in the graph $G$. 



\section{Methodology}
\label{sec::model}


\subsection{Overview}

The paper proposes the \method{} for next POI recommendation, which jointly leverages geographical and sequential influences for CTR prediction on the target POI. We argue that geographical influence and sequential influence are two of the major driving forces in POI recommendation. However, existing methods typically fail to explore topological geographical influences among POIs and meanwhile show the inability to model high-order sequential substructures.

To better explore high-order interaction information in user behavior sequences, \method{} integrates the key factors of geographical influence and sequential influence. More specifically, \method{} has a geographical module to capture the topological geographical influences and a sequential module to explore high-order sequential substructures. To effectively train both modules, a consistency learning framework is introduced to incorporate geographical and sequential information jointly. The two modules are optimized by distilling the knowledge from each other and providing mutual supervision signals. The overall framework of \method{} is shown in Fig.~\ref{fig:framework}.

\subsection{The Geographical Module}

\subsubsection{Construction of Geographical Graph}
In view of the fact that nearby things are more related than distant things. Technically, we first calculate the physical distance between POIs from their unique location and construct a geographical graph $\mathcal{G}_g=\{\mathcal{V}, \mathcal{E}_g, A_g\}$, where edge $e_g=(v_i,v_j)\in\mathcal{E}_g$ indicates the distance between $v_i$ and $v_j$ less than a specific distance threshold $\Delta d$. The edge weight matrix $A_g(i,j)$ means the physical distance between POI $v_i$ and $v_j$ in kilometers. The construction of $\mathcal{G}_g$ depicts the prior knowledge of user preferences, i.e., users prefer to visit neighboring POIs with close physical distance.

\smallskip
\subsubsection{Message Passing Neural Network for Geographical Graph}

Due to the remarkable success of GNNs in capturing high-order relations on the graph topology, given the constructed geographical graph $\mathcal{G}_g$, we further leverage message passing neural networks~\cite{kipf2017semi} to explore complex and topological distance-based geographical influences.

Technically, for a pair of neighboring POIs $v_i, v_j$, the message from $v_i$ to $v_j$ on $l$-th layer is defined as:
\begin{equation}
\label{eqn:message_construction}
    m_{j\leftarrow i}^{(l)}=\frac{1}{\sqrt{|\mathcal{N}_i||\mathcal{N}_j|}}w(d_{ij})W^{(l)}h_i^{(l-1)},
\end{equation}
where $W^{(l)}\in\mathbb{R}^{D\times D}$ are trainable weight matrices. $h_i^{(l-1)}$ represents POI representations from the $(l-1)$-th layer of GNNs. $h_i^{(0)}$ is often initialized as POI's embedding $X$ (i.e., $h_i^{(0)}=x_i, \forall v_i\in\mathcal{V}$). To better explore the geographical influence in $\mathcal{G}_g$, the message is desired to reveal the distance influence between POIs, we hence introduce distance kernel $w(d_{ij})=e^{-d_{ij}^2}$ which decays exponentially as $v_i$ and $v_j$ become farther apart, where $d_{ij}$ represents the distance between $v_i$ and $v_j$. Intuitively, POIs which are closer tend to share more similarities than distant ones. Additionally, here we use the same graph convolutional rule based on symmetric normalized Laplacian as GCN~\cite{kipf2017semi}, where $\mathcal{N}_i$ and $\mathcal{N}_j$ denote the first-hop neighbors for $v_i$ and $v_j$ on $\mathcal{G}_g$ respectively.

To generate effective POI representations based on message-passing mechanism, the process of neighbor aggregation and information update is defined as:
\begin{equation}
    h_j^{(l)}=\text{LeakyReLU}(m_{j\leftarrow j}^{(l)}+\sum_{i\in\mathcal{N}_{v_j}}m_{j\leftarrow i}^{(l)}),
\end{equation}
where $\mathcal{N}_{v_j}$ are neighbors to node $v_j$, and $m_{j\leftarrow j}^{(l)}=W^{(l)}h_j^{(l-1)}$ is the message generated by self-connection of node $v_j$.

The message-passing phase runs for $L$ layers, and we use the representations of the $L$-th layer as the geographical encoding of all the POIs in $\mathcal{V}$:
\begin{equation}
H_g=[h_1^{(L)};h_2^{(L)};...;h_{|\mathcal{V}|}^{(L)}].
\end{equation}

In practice, given a user $u$ with check-in behavior sequence $s_u$, his/her geographical encoding $H_{g,u}$ is the list of the geographical representations of all the POIs that appear in $s_u$. 
Here we assume that the geographical preferences of users can be captured by aggregating the high-order geographical POI representations in his/her check-in behaviors.  

To produce effective semantic representations of user geographical preferences for capturing topological geographical influences, we propose to leverage a multi-head self-attention mechanism to better aggregate different POIs. The semantic geographical representations can be formulated as:
\begin{equation}
\begin{gathered}
    \alpha_{i}^r = \frac{(Q^r\cdot h^{(L)}_i)^{\top}(K^r\cdot h^{(L)}_i)}{\sqrt{d/R}}, \hat{\alpha}_{i}^r = \frac{\exp{\alpha_{i}^r}}{\sum_{j=1}^{|s_u|}\exp{\alpha_{j}^r}}\\
    e_{g,u} = \overset{R}{\underset{r=1}{\Vert}}\sum_{i=1}^{|s_u|}\hat{\alpha}_{i}^r V^r\cdot h_i^{(L)},
    \label{eqn:self_attention}
\end{gathered}
\end{equation}
where $Q^r,K^r$ and $V^r \in \mathbb{R}^{(d/R)\times d}$ are query, key and value transformation matrices, respectively. $\hat{\alpha}_{i}^r$ is viewed as the importance of the $i$-th representation. $R$ is the number of heads, $\Vert$ is the concatenation operator. Finally, we aggregate different POIs representations of the whole geographical graph to obtain a semantic geographical representation $e_{g,u}$.

\subsection{The Sequential Module}

\subsubsection{Construction of Sequential Graph}

In addition to geographical graph influencing user behaviors, intrinsic characteristics of POIs are critical factors that affect user preferences. We therefore construct a sequential graph $\mathcal{G}_{s,u}=\{\mathcal{V}_{s,u},\mathcal{E}_{s,u}\}$ given a user $u$ and his/her historical check-in sequence $s_u$. Each edge $e_{s,u}=\left\langle v_i,v_j\right\rangle\in\mathcal{G}_{s,u}$ represents the user's successive check-ins from $v_i$ to $v_j$. The construction of $\mathcal{G}_{s,u}$ describes the sequential dependencies of users' check-in behaviors.

\smallskip
\subsubsection{Graph Kernel Neural Network for Sequential Graph}
Owing to the fact that behavior sequences of the users contain not only the information about their successive preferences on POIs, but also several possible regular routines, which implies the path dependency and some specific substructure patterns (i.e., `triangle' or a `square'). Inspired by the superiority of graph kernels~\cite{gartner2003graph,kashima2003marginalized,ju2022kgnn,feng2022kergnns} in capturing high-order substructures (e.g., random walk~\cite{kashima2003marginalized,nikolentzos2020random}), we propose to leverage a random walk graph kernel to explicitly explore sequential influences and path dependencies on $\mathcal{G}_{s,u}$. 


To efficiently compute the random walk kernels, here we introduce the direct product graph defined as below.

\smallskip
\noindent\textbf{Definition 4: (Direct Product Graph)}
Given two graphs $G=(\mathcal{V},\mathcal{E})$ and $G'=(\mathcal{V'},\mathcal{E'})$, define the graph direct product graph $G_\times = (\mathcal{V_\times}, \mathcal{E_\times}) $, where \(\mathcal{V_\times}=\left\{\left(v, v^{\prime}\right): v \in \mathcal{V} \wedge v^{\prime} \in \mathcal{V'}\right\}\) and \(\mathcal{E_\times}=\left\{\left\{\left(v, v^{\prime}\right),\left(u, u^{\prime}\right)\right\}:\{v, u\} \in \mathcal{E} \wedge\left\{v^{\prime}, u^{\prime}\right\} \in \mathcal{E'}\right\}\).

Note that a random walk on $G_\times$ can be interpreted as a simultaneous walk on graphs $G$ and $G'$~\cite{vishwanathan2010graph}.  The $p$-step ($p \in \mathbb{N}$) random walk kernel between $G$ and $G'$ that counts all simultaneous random walks is thus defined as:
\begin{equation}
k^{(p)}\left(G, G^{\prime}\right)= \mathbf{e}^{\top} \mathbf{A}_{\times}^{p}\mathbf{e},
\end{equation}
where $\mathbf{e}$ is an all-one vector,  $\mathbf{A}_\times$ is adjacency matrix of $G_\times$.

Motivated by filters in convolutional neural networks, we introduce a set of trainable graph filters to extract the structural information (i.e., sequential patterns) of the sequential graph.

\smallskip
\noindent\textbf{Definition 5: (Graph Filter)}
The $i$-th graph filter $G'_i$ is a graph with $n_{i}$ nodes and we parametrize each graph filter with a trainable adjacency matrix $A_{i}\in\mathbb{R}^{n_{i}\times n_{i}}$.

These graph filters are expected to learn high-order substructures (e.g., random walk, paths) that capture high-order sequential patterns of user preferences. 

Inspired by the fact that the random walk kernels quantify the similarity of two graphs based on the number of common walks in the two graphs~\cite{kashima2003marginalized},
we compare each sequential graph of the user against graph filters with a differentiable function from the random walk kernel. 

Then, given different random walk lengths $\mathcal{P} = \{0,\dots, P \}$ and a set of graph filters $\mathbf{\mathcal{G}}_h = \{G'_1, \dots, G'_N\}$, we can construct a matrix $\mathbf{H}\in \mathbb{R}^{N\times (P+1)}$ where $\mathbf{H}_{ij} = k^{(j-1)}(G_{s,u}, G'_i)$ for each sequential graph $G_{s,u}$. Finally, the matrix $\mathbf{H}$ is flattened and fed into a fully-connected layer to produce a semantic sequential representation denoted as $e_{s,u}$:
\begin{equation}
\begin{split}
   e_{s,u} = \mathrm{CONCAT}\left(\mathbf{H}_{i,j} \,\Big\vert\, \forall i \in N, j\in\mathcal{P} \right).
\end{split}
\label{eqn:kernel}
\end{equation}

In this way, we are capable of exploring high-order sequential substructures in the user-aware sequential graph via a graph kernel neural network to capture user preferences. By visualizing the graph filters in experiments, we further show the superiority in capturing high-order sequential substructures.

\subsection{Joint Optimization Framework}
In this section, we discuss how to integrate two modules jointly to explore topological geographical influences and meanwhile model high-order sequential substructures. 

After encoding the two constructed graphs from the user's check-in sequences via two modules. How to integrate this two complementary information is a crucial concern. Due to the difference in behavior semantics extracted from the sequential module and geographical module respectively, it is necessary to need a way to combine the advantages of both sequential and geographical influences. However, due to the data sparsity of check-in behaviors, directly aligning the semantic representations may lead to sub-optimal performance.
To alleviate the issue, we propose to enhance each user behavior sequence via comparing its similarities to other behavior sequences in embedding spaces of two modules. In this way, the embedding spaces from two modules can be matched in a soft manner and achieve a smooth consistency.

Technically, we begin to sample a range of the behavior sequences $\{s_{u_1},s_{u_2},\dots, s_{u_M}\}$ as anchors and then leverage a memory bank to store them. Then, we embed them with both the geographical module and the sequential module, resulting in semantic geographical representations and semantic sequential representations, respectively. Unfortunately, we find that there is a dilemma, where we need a large number of anchors with substantial variability to cover the vicinity of the whole dataset, while processing such massive sequences simultaneously is computationally costly. To circumvent this, the memory bank is maintained as a queue that is dynamically modified by anchor sequences from the most recent iterations.
 
In particular, we generate two semantic representations using both the geographical module and the sequential module for each sequence. Then the pairwise similarity between its embedding with all the anchor embeddings in both embedding spaces. The similarity distribution between each sequence and all anchors for the geographical module is written as:
\begin{equation}
p_{g,u}^m=\frac{\Phi(e_{g,u}, e_{g,u_m})}{\sum_{m'=1}^{M} \Phi(e_{g,u}, e_{g,u_{m'}})},
\end{equation}
where $\Phi(\cdot, \cdot) = \exp \left( \cos(\cdot, \cdot) / \tau\right)$ is implemented by the popular exponential temperature-scaled cosine metric to measure the similarity and 
$\tau$ denotes the temperature set to $0.5$ following \cite{chen2020simple}. $\cos(\cdot,\cdot)$ denotes the cosine metric. Similarly, we can produce the similarity for the sequential module as:
\begin{equation}
p_{s,u}^m=\frac{\Phi(e_{s,u}, e_{s,u_m})}{\sum_{m'=1}^{M} \Phi(e_{s,u}, e_{s,u_{m'}})},
\end{equation}

In the end, we attempt to maximize the consistency between two derived distributions, i.e., $p_{s,u}=[p_{s,u}^1, \dots, p_{s,u}^M]$ and $p_{g,u}=[p_{g,u}^1, \dots, p_{g,u}^M]$ , which can allow the knowledge from two modules to communicate with each other for harmonious structures in both embedding spaces. The final consistency loss is formulated as follows:
\begin{equation}
\label{eqn:consistency}
\mathcal{L}_{\text{con}}=\frac{1}{|\mathcal{U}|} \sum_{u \in \mathcal{U}} \frac{1}{2}\left(D(p_{s,u} \| p_{g,u})+ D(p_{g,u} \| p_{s,u})\right),
\end{equation}
where $D(\cdot\|\cdot)$ is implemented using the Kullback-Leibler divergence to measure the difference between two distributions.

\smallskip
\noindent\textbf{Comparison with Contrastive Learning.} Although our consistency learning contains similar parts to contrastive learning, they have the following differences: (i) Contrastive learning involves two views by data augmentation while our consistency learning involves two modules to mine information from complementary views; (ii) Contrastive learning usually maximizes the mutual information while our consistency learning attempts to minimize the distribution difference; (iii) Contrastive learning employs a ``hard" manner to achieve alignment from two views in the same embedding space while our consistency learning allows the discordance of two spaces and employs a ``soft" manner to achieve semantics consistency.

\subsection{Training Algorithm}

To better improve the CTR prediction~\cite{zhou2018deep,zhou2019deep}, we integrate the semantic representations derived from two modules according to the current target POI $v_t$. Specifically, we concatenate the semantic geographical representation $e_{g,u}$ and the semantic sequential representation $e_{s,u}$, along with target $v_t$'s geographical representation $h_{t}$. Afterward, a two-layer multi-layer perceptron (MLP) and a sigmoid function $\sigma$ are applied to predict the probability of target POI $v_t$ being clicked:
\begin{equation}
\label{eqn:pred}
    \hat{y}=\sigma(\text{MLP}([e_{g,u},e_{s,u},h_{t}])).
\end{equation}

Given the ground-truth click-through rate $y \in \{0,1\}$, the objective function of supervised CTR prediction is evaluated via binary cross-entropy loss defined as:
\begin{equation}
    \mathcal{L}_{\text{rec}}=-\sum_{(u,v_t)}y\log\hat{y}+(1-y)\log(1-\hat{y}).
\label{eqn:ctr_prediction}
\end{equation}
The self-supervised consistency learning proceeds as described in Eq.~\ref{eqn:consistency}, acting as an auxiliary task that is unified into a joint learning objective to enhance the next POI recommendation. The overall objective of the joint learning can be written as:
\begin{equation}
    \mathcal{L}=\mathcal{L}_{\text{rec}}+\beta * \mathcal{L}_{\text{con}},
\label{eqn:overall_loss}
\end{equation}
where $\beta$ is the tuning parameter used to control the magnitude of self-supervised consistency learning and supervised CTR prediction, which sets $\beta=0.01$ in practice. The overall process of \method{} is presented in Algorithm \ref{alg:1}.

\begin{algorithm2e}[t]
\LinesNumberedHidden
\newcounter{algoline}
\newcommand\Numberline{\refstepcounter{algoline}\nlset{\thealgoline}}
\SetNlSkip{0em}
\SetAlgoNlRelativeSize{0}
\caption{Learning Algorithm of \method{}}
\label{alg:1}
\KwIn{POI set $\mathcal{V}$}
\SetKwInput{kwInit}{Initialize}
\kwInit{Distance matrix $A_g$; Model parameters}
Construct $\mathcal{G}_{g}$ based on $A_g$\;
\While{not converged}{
Sample POI sequence $s_u=[v_1,v_2,...,v_{|s_u|}]$ and the corresponding target POI $v_t$\;
Construct geographical graph $\mathcal{G}_{g,u}$ and sequential graph $\mathcal{G}_{s,u}$\;
\tcc{Eq.\ref{eqn:self_attention}, Eq.\ref{eqn:kernel}}
Obtain semantic representations $e_{g,u}$, $e_{s,u}$\;
\tcc{Eq.\ref{eqn:consistency}, Eq.\ref{eqn:ctr_prediction}}
Compute objective function $\mathcal{L}_{\text{rec}},\mathcal{L}_{\text{con}}$\;
\tcc{Eq.\ref{eqn:overall_loss}}
Optimize framework via $\mathcal{L}=\mathcal{L}_{\text{rec}}+\beta*\mathcal{L}_{\text{con}}$
}
\end{algorithm2e}
\section{Experiment}
\label{sec::experiment}
In this section, we conduct extensive experiments on two real-world POI recommendation datasets to investigate the effectiveness and robustness of the proposed \method{}.
We attempt to answer the following research questions:
\begin{itemize}
    \item \textbf{RQ1}: How well does our \method{} perform against the baseline models on CTR prediction? Does consistency learning help under cold-start settings?
    \item \textbf{RQ2}: How does each part of the model affect the recommendation? How do hyper-parameters influence the model performance?
    \item \textbf{RQ3}: Does \method{} model topological geographical influences of POIs via the geographical module? What high-order substructures are captured by the sequential module? Can two modules visualize the actual case?
\end{itemize}

\begin{table}
\centering
\caption{Basic statistics for our two datasets.}
\label{tab:1}
\label{tab:statics}
\setlength{\tabcolsep}{3pt}
\begin{tabular}{c|cccc} 
\toprule 
\textbf{Dataset} & \#User & \#POI & Interactions & Avg.SeqLen \\
\midrule 
Tokyo & 2,293 & 61,858 & 573,703 & 250.20 \\
\midrule
New York & 1,083 & 38,333 & 227,428 & 210.00 \\
\bottomrule 
\end{tabular}
\end{table}

\subsection{Experiment Setup}


\paragraph{Datasets}
We conduct experiments on two industrial POI recommendation datasets~\cite{yang2014modeling}, which are drawn from the users' check-in histories on \textbf{Foursquare}\footnote{{\url{https://sites.google.com/site/yangdingqi/home/foursquare-dataset}}}. The datasets contain user check-ins from two cities, namely \textbf{Foursquare Tokyo} (TKY) and \textbf{New York} (NYC). The check-in records are collected in two cities from 12 April 2012 to 16 February 2013. The statistics of datasets are shown in Table~\ref{tab:1}.

We sort each user's check-in sequence in chronological order. The last POI visited by each user is reserved as the evaluation set, with the rest of the POIs in the sequence serving as training data. The evaluation set is randomly split into equal-sized test and validation sets. 

\paragraph{Baselines}

We compare the proposed \method{} with a wide range of baselines. To ensure the diversity, we select the current state-of-the-art baselines from three categories: (1) Traditional sequential-based models for CTR prediction; (2) Graph-based models that leverage GNNs for recommendation; (3) POI recommendation methods that incorporate the geographical influence when making predictions.

\smallskip\noindent\textbf{(i) Sequential-based models:}
\begin{itemize}
    \item \textbf{DIN}\cite{zhou2018deep}: It is one of the most classical sequential-based models for CTR prediction that proposes target attention via outer product between the target item and user history.
    \item \textbf{DIEN}\cite{zhou2019deep}: This model is a variant of DIN that leverages the Gated Recurrent Unit (GRU) to model the evolving history of user interests. 
\end{itemize}

\smallskip\noindent\textbf{(ii) GNN-based models:}
\begin{itemize}
    \item \textbf{SR-GNN}\cite{wu2019session}: It is a session recommendation model that adopts Gated Graph Neural Networks (GGNN) to propagate information on the transmission graph of items.
    \item \textbf{NGCF}\cite{wang2019neural}: It is a graph-based collaborate filtering recommendation model. NGCF propagates on user-item bipartite graph to capture collaborative signals.
    \item \textbf{LightGCN}\cite{he2020lightgcn}: It is a variant of NGCF, which achieves superior results for graph-based recommendation.
\end{itemize}

\smallskip\noindent\textbf{(iii) POI recommendation models:}
\begin{itemize}
    \item \textbf{GeoIE}\cite{wang2018exploiting}: It models the geographical influence of the historical POIs on the current target with two sets of locational embeddings.
    \item \textbf{LSTPM}\cite{sun2020go}: It proposes geo-dilated LSTM  structure to expand LSTM to capture both temporal and spatial similarities between POIs. 
    \item \textbf{GSTN}\cite{wang2022graph}: It models the transmission between POIs on a graph. GSTN represents the POIs on graphs by minimizing the difference between conditional transmission probability and empirical probability.
\end{itemize}

We adopt AUC and Logloss as evaluation metrics to evaluate the mentioned models for CTR prediction. 
The embedding size of hidden layers of all models is fixed to $64$. For the \method{}, the consistency weight $\beta$ is set to $0.01$. The threshold distance $\Delta d$ when constructing the geographical POI graph is set to $0.5$ km. 
We summarize the performance of each model by the average over three randomly initialized experiments.

\begin{table}
\centering
\caption{Performance comparison of all compared methods.}
\label{tab:2}
\begin{tabularx}{\linewidth}{cYYYY}
\toprule 
\multirow{2}{*}{Model} & \multicolumn{2}{c}{Tokyo} & \multicolumn{2}{c}{NYC}\\ 
\cmidrule[0.5pt](lr){2-3}\cmidrule[0.5pt](lr){4-5}
& AUC$\uparrow$ & Logloss$\downarrow$ & AUC & Logloss\\
\midrule 
DIN & 0.8623 & 0.4174 & 0.8015 & 0.5203 \\
DIEN & 0.8643 & 0.4544 & 0.8062 & 0.5582 \\
\midrule
SR-GNN & 0.8867 & 0.4481 & 0.8491 & 0.5262 \\
NGCF & 0.8899 & 0.4436 & 0.8461 & 0.4788 \\
LightGCN & 0.8998 & 0.4348 & 0.8729 & 0.4815 \\
\midrule
GeoIE & 0.9083 & 0.4203 & 0.8769 & 0.4810 \\
LSTPM & 0.8745 & 0.4374 & 0.8568 & 0.4821 \\
GSTN & 0.8906 & 0.4312 & 0.8521 & 0.5323 \\
\midrule
Ours & \textbf{0.9387} & \textbf{0.3147} & \textbf{0.9189} & \textbf{0.3620}\\
\bottomrule 
\end{tabularx}
\end{table}

\subsection{Performance Comparison (RQ1)}
\paragraph{Overall Comparison}
We first train and evaluate the proposed model and baselines on two full datasets. From the results shown in table \ref{tab:2}, we can observe that:
\begin{itemize}[leftmargin=*]
    \item Superiority of geographical influences: It can be seen that there is a clear advantage for models that leverage geographical information to assist recommendation (GeoIE, LSTPM, GSTN and our model) over other baseline methods. These results show that it is essential to consider the geographical influences of different POIs when recommending the next-to-visit POI. Specifically, GeoIE achieves the best performance among all baselines, which shows the importance of modeling locational similarities between POIs.
    \item Superiority of Graph Neural Networks: There is a common improvement made by GNN-based models (SR-GNN, NGCF, LightGCN and \method{}) compared with traditional sequential-based models (DIN and DIEN). The results demonstrate the strong capability of GNNs to capture the high-order connectivity between nodes of different POIs. While sequential-based methods are faced with challenges brought by long sequences, GNN-based models function well with the help of rich neighborhood similarities.
    \item Overall, our proposed \method{} outperforms all the baseline methods significantly on both datasets. To be specific, the testing AUC is improved by more than 3.3\% and 4.5\%, and the testing Logloss is improved by 23\% and 24\% on \textbf{Tokyo} and \textbf{NYC} respectively over the best of the baselines. \method{} achieves the state-of-the-art performance, which proves the effectiveness of combining two GNN modules to recommend the next-to-visit POIs.
\end{itemize}

\begin{figure}
\centering
\subfigure[AUC on \textbf{Tokyo}]{
\begin{minipage}{0.49\linewidth}
    \includegraphics[width=\linewidth]{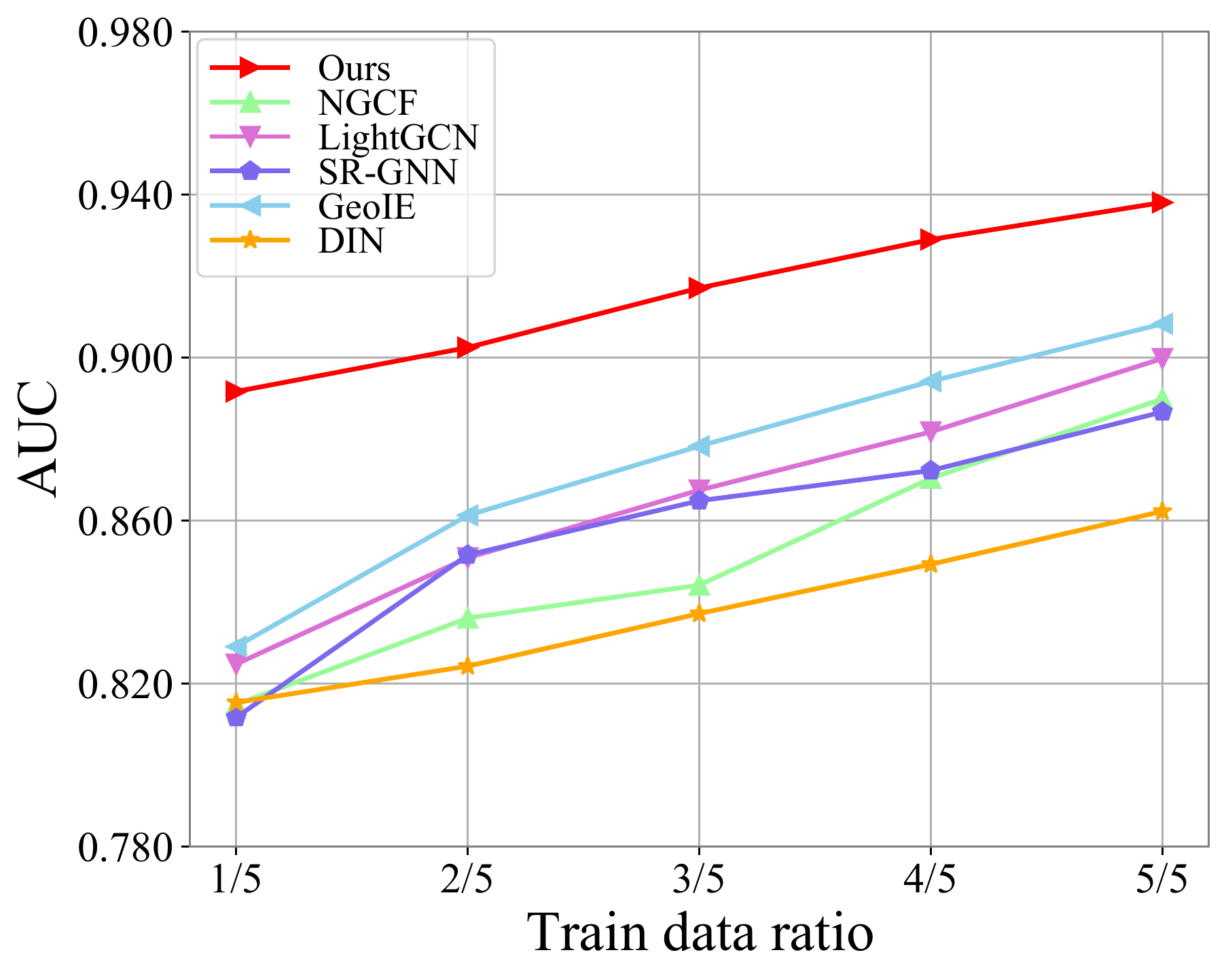}
    \label{fig:colda}
\end{minipage}
}%
\subfigure[AUC on \textbf{NYC}]{
\begin{minipage}{0.49\linewidth}
    \includegraphics[width=\linewidth]{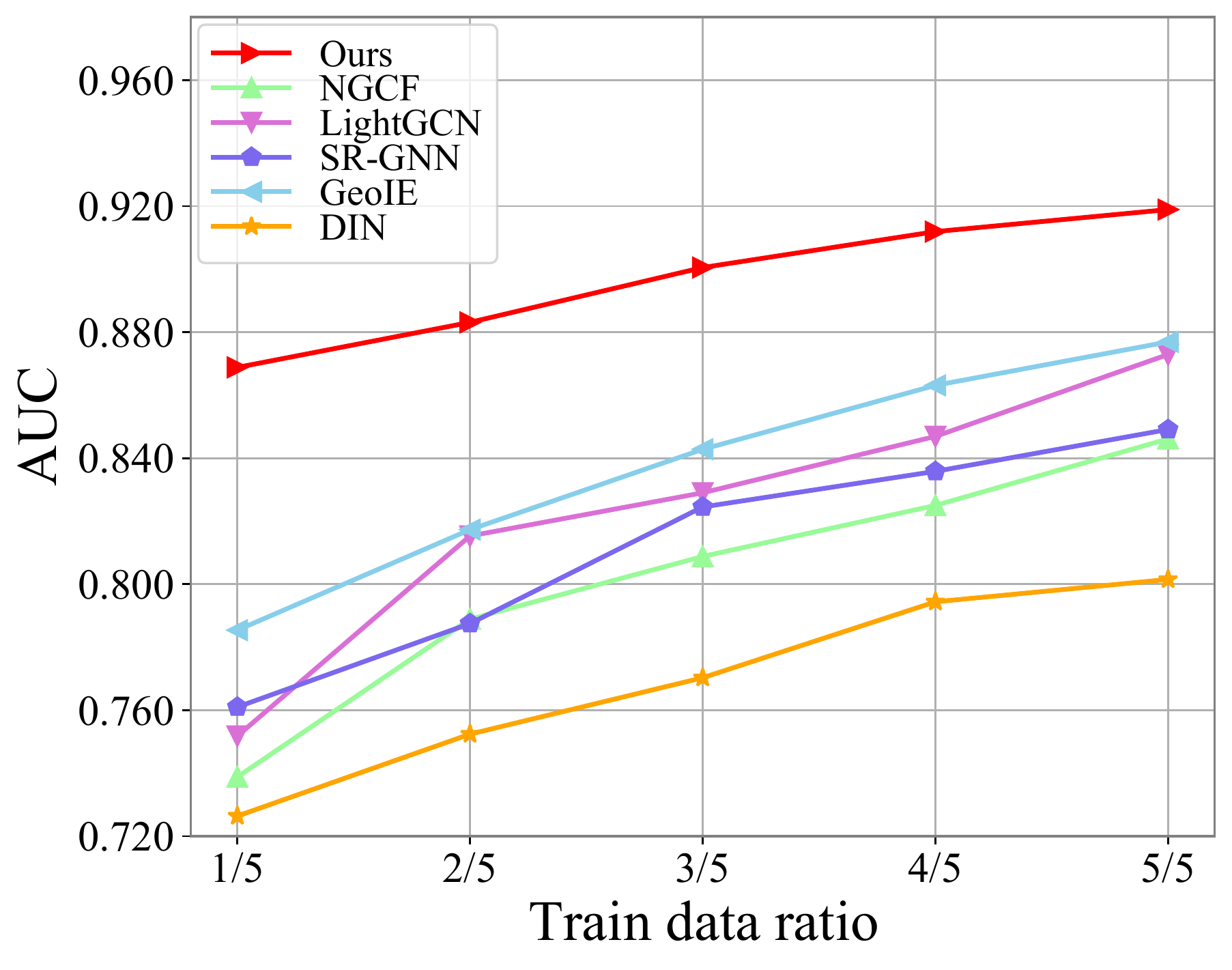}
    \label{fig:coldb}
\end{minipage}
}

\subfigure[Logloss on \textbf{NYC}]{
\begin{minipage}{0.49\linewidth}
    \includegraphics[width=\linewidth]{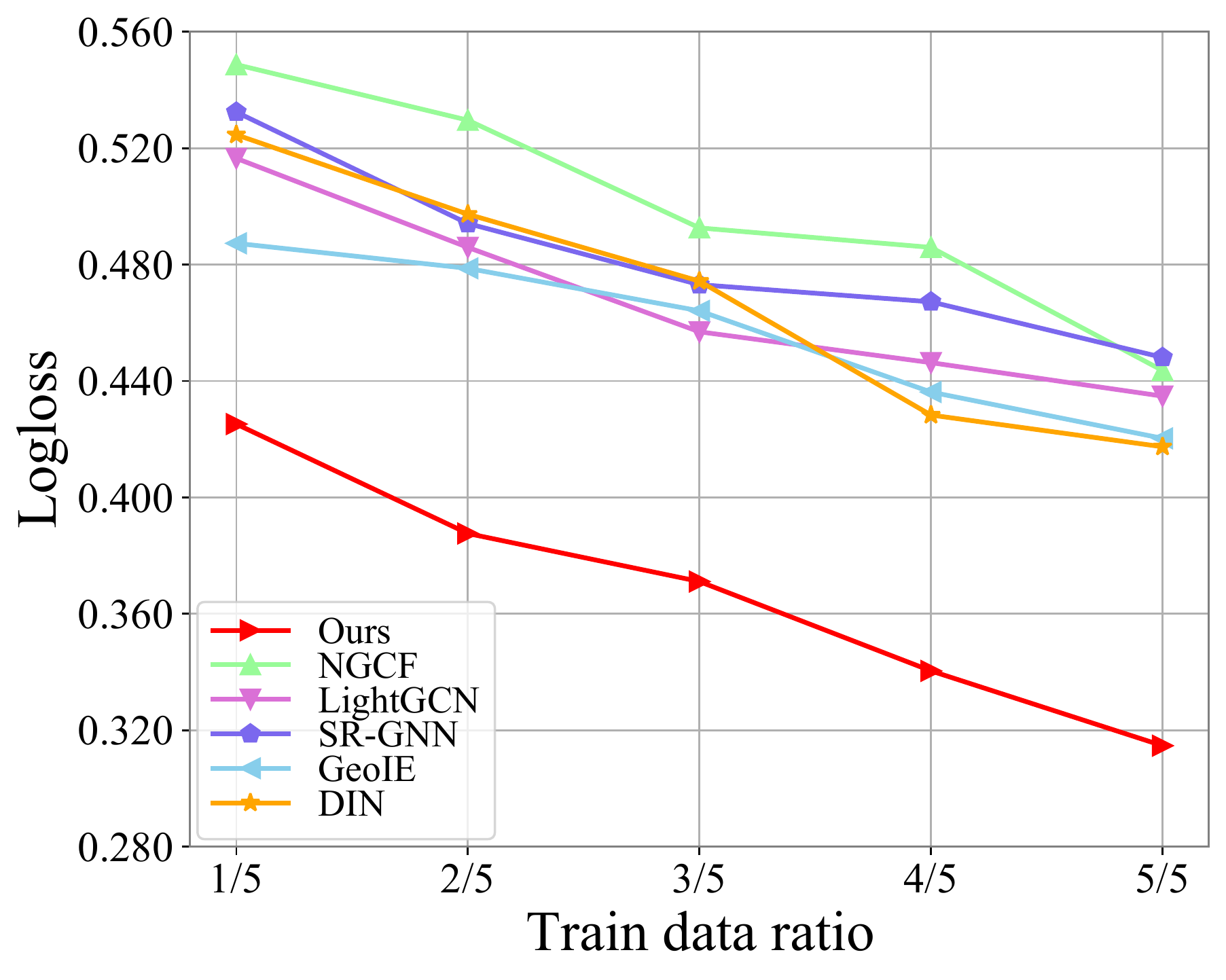}
    \label{fig:coldc}
\end{minipage}
}%
\subfigure[Logloss on \textbf{NYC}]{
\begin{minipage}{0.49\linewidth}
    \includegraphics[width=\linewidth]{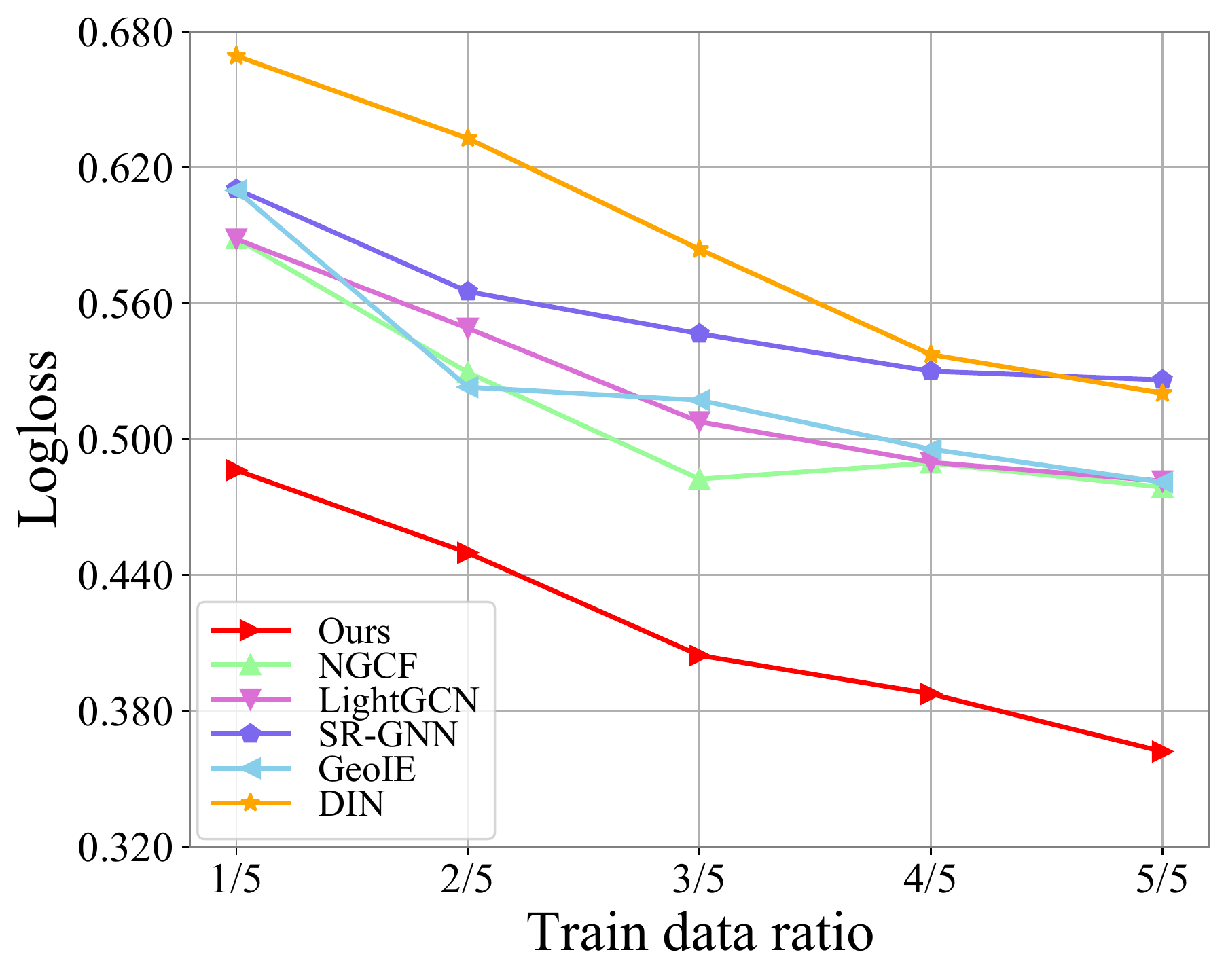}
    \label{fig:coldd}
\end{minipage}
}
\caption{Performance comparison under cold-start settings. The data sparsity of the train set is ranged from $20\%$ of the data (one fold of all splits) to $100\%$ (whole train set).}
\label{fig:cold}
\end{figure}

\paragraph{Cold-start Settings}
Due to the issue of data sparsity, i.e. cold-start problem, which remains a tough challenge in real-world recommendations, we want to figure out whether the proposed \method{} can alleviate this effect with the knowledge of the sequential and geographical semantic representations. Towards this end, we divide the full train set into five folds equally and take one to five folds each time, corresponding to 20\%, 40\%, 60\%, 80\%, and full train set, respectively. All models are trained on these subsets of the train set to simulate the cold-start settings with varying degrees of data sparsity. The results illustrated in Fig.~\ref{fig:cold} show that:
\begin{itemize}[leftmargin=*]
    \item The data sparsity problem has made the performance drop severely for all methods. Surprisingly, \method{} consistently outperforms all baselines under different data sparsity settings, which demonstrates the robustness and reliability of our joint framework. Moreover, we can see that our model achieves high enough performance when the training data is extremely sparse (from 20\% to 40\%), which pushes away the other baselines, sufficiently showing the superiority of knowledge intra-communication between two modules.
    \item Generally, models that leverage geographical information (GeoIE, LSTPM, and \method{}) perform better on the data with high sparsity. Maybe the reason is that when the interaction information is partly missing, the model could still fully utilize the locational feature of all POIs to make a recommendation. This observation indicates that geographical influence plays an essential role when faced with challenges brought by data sparsity.
\end{itemize}
\subsection{Ablation and Hyper-parameter Study (RQ2)}
To further investigate the contribution of each part of the proposed \method{}, we conduct the ablation study and hyper-parameter study, which reveal the actual mechanism behind the whole process when making recommendations.

\smallskip
\paragraph{Ablation Study of Two GNN Modules}

\begin{figure}
\centering
\subfigure[Results on \textbf{Tokyo}]{
\begin{minipage}{0.49\linewidth}
    \includegraphics[width=\linewidth]{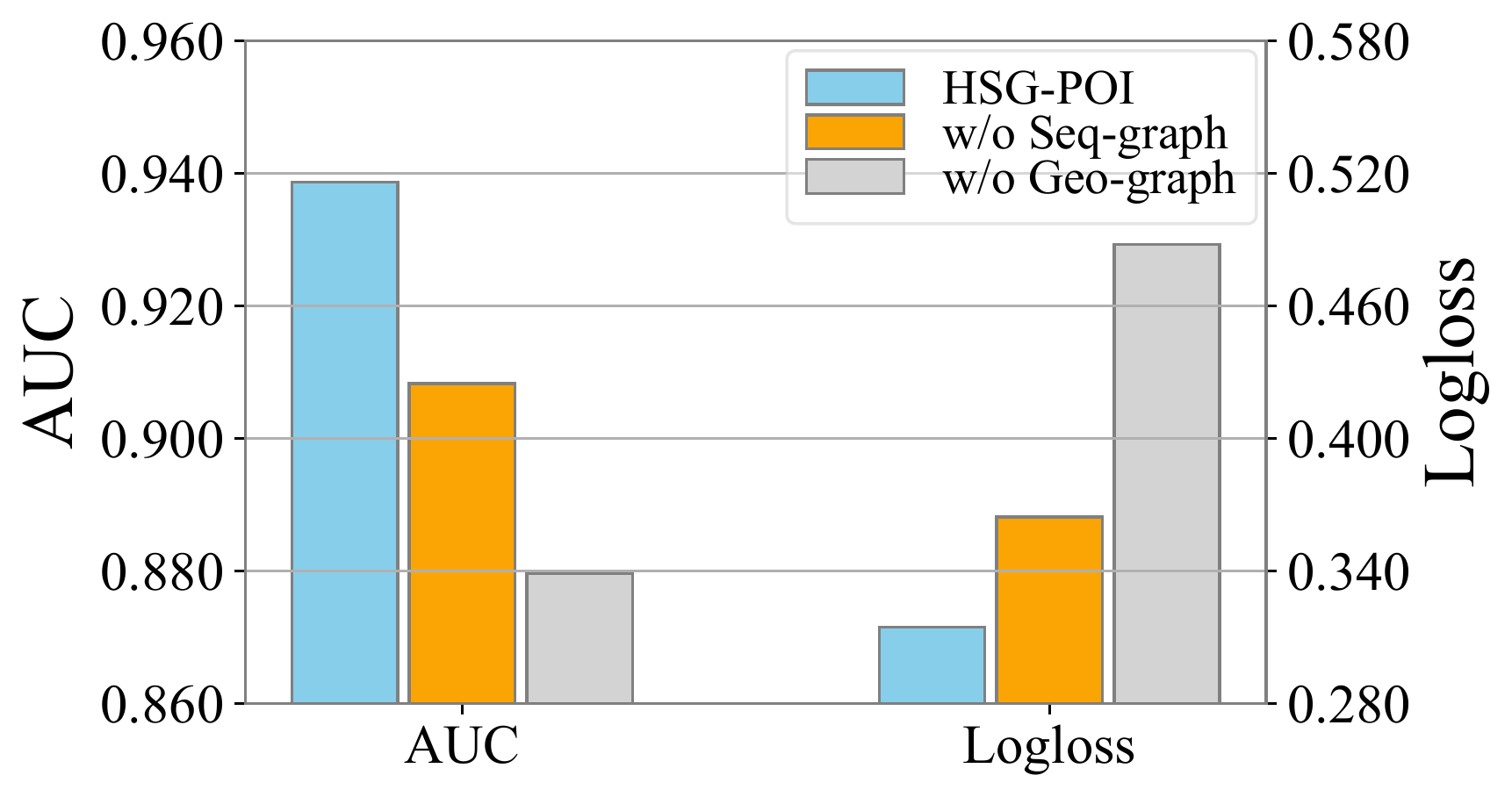}
    \label{fig:singlea}
\end{minipage}
}%
\subfigure[Results on \textbf{NYC}]{
\begin{minipage}{0.49\linewidth}
    \includegraphics[width=\linewidth]{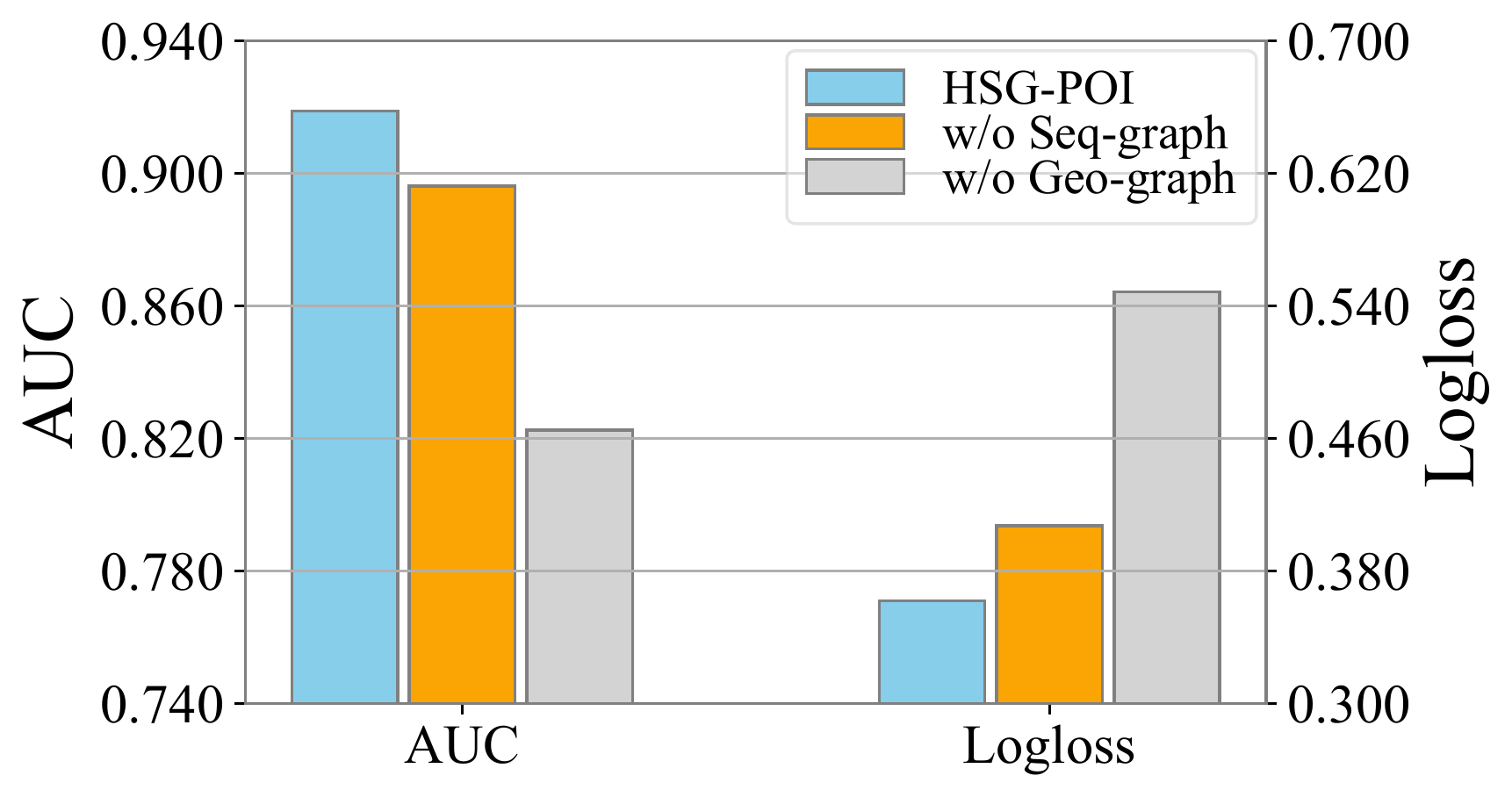}
    \label{fig:singleb}
\end{minipage}
}
\caption{Performance of \method{} on two datasets without sequential graph (w/o Seq-graph) and without geographical graph (w/o Geo-graph).}
\label{fig:single}
\end{figure}

The \method{} extracts semantic features from two different graphs. It is necessary to figure out how the two GNN modules cooperate. We conduct ablation studies against these two modules by removing one of the GNN modules each time.
To be specific, we only train one of the modules via the supervised signals, which means the constraints of consistency learning are also removed. The results summarized in Fig.~\ref{fig:single} reveal that:
\begin{itemize}[leftmargin=*]
    \item Overall, the results of \method{} are consistently better than all the other two variants, indicating that both the geographical and the sequential modules are effective for recommendation. It further verifies that geographical influence and sequential influence are two of the key factors in POI recommendation, which aligns with our expectations.
    \item Geographical influences are more essential for next POI recommendation. Specifically, the model would suffer a more severe performance decline when the geographical module is removed (w/o Geo-graph), compared with the model with the sequential module removed (w/o Seq-graph). These results indicate that geographical influences play a more vital role when deciding where to visit next.
\end{itemize}

\begin{figure}
\centering
\subfigure[Different layer numbers on \textbf{Tokyo}]{
\begin{minipage}{0.49\linewidth}
    \includegraphics[width=\linewidth]{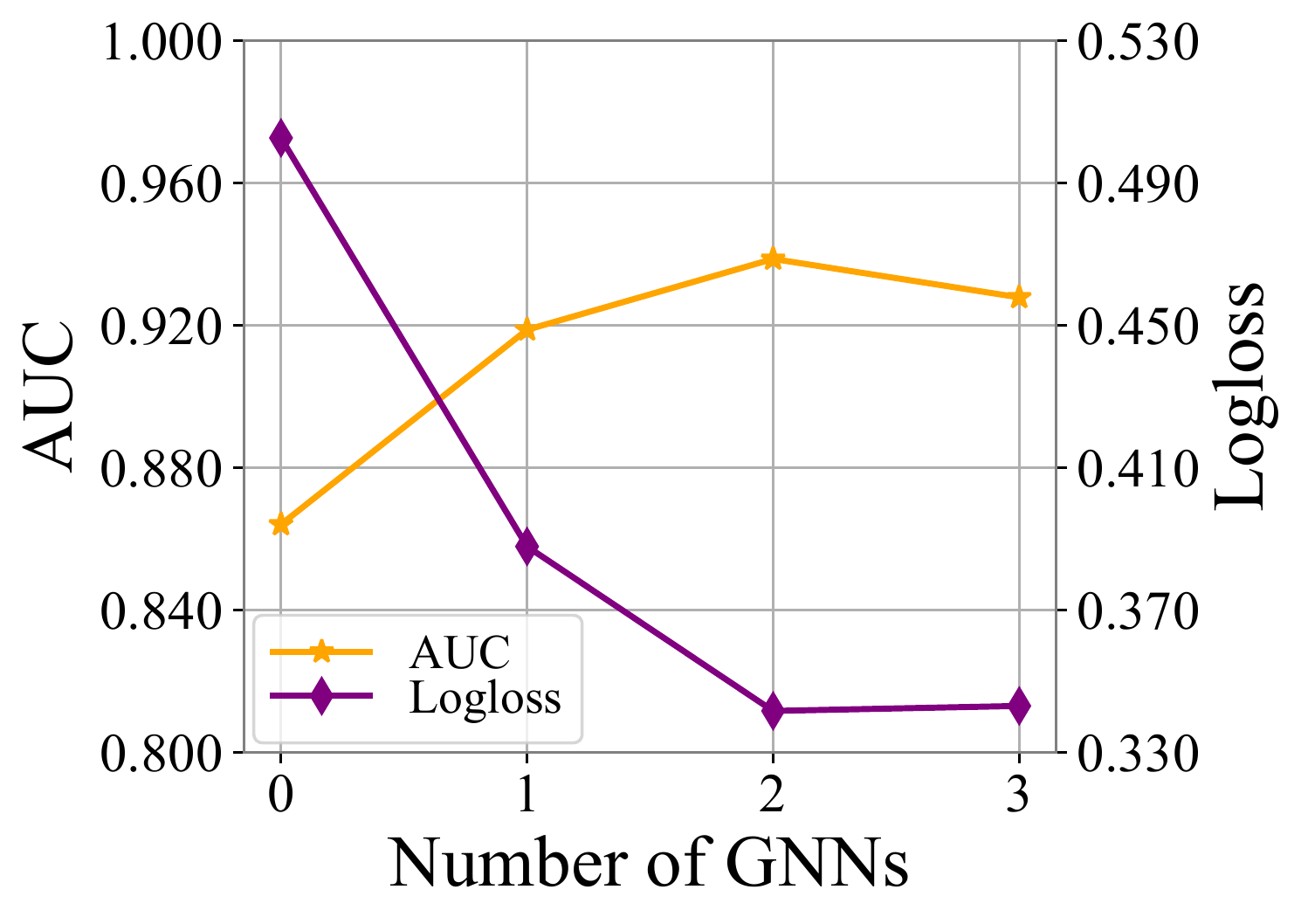}
    \label{fig:parama}
\end{minipage}
}%
\subfigure[Different layer numbers on \textbf{NYC}]{
\begin{minipage}{0.49\linewidth}
    \includegraphics[width=\linewidth]{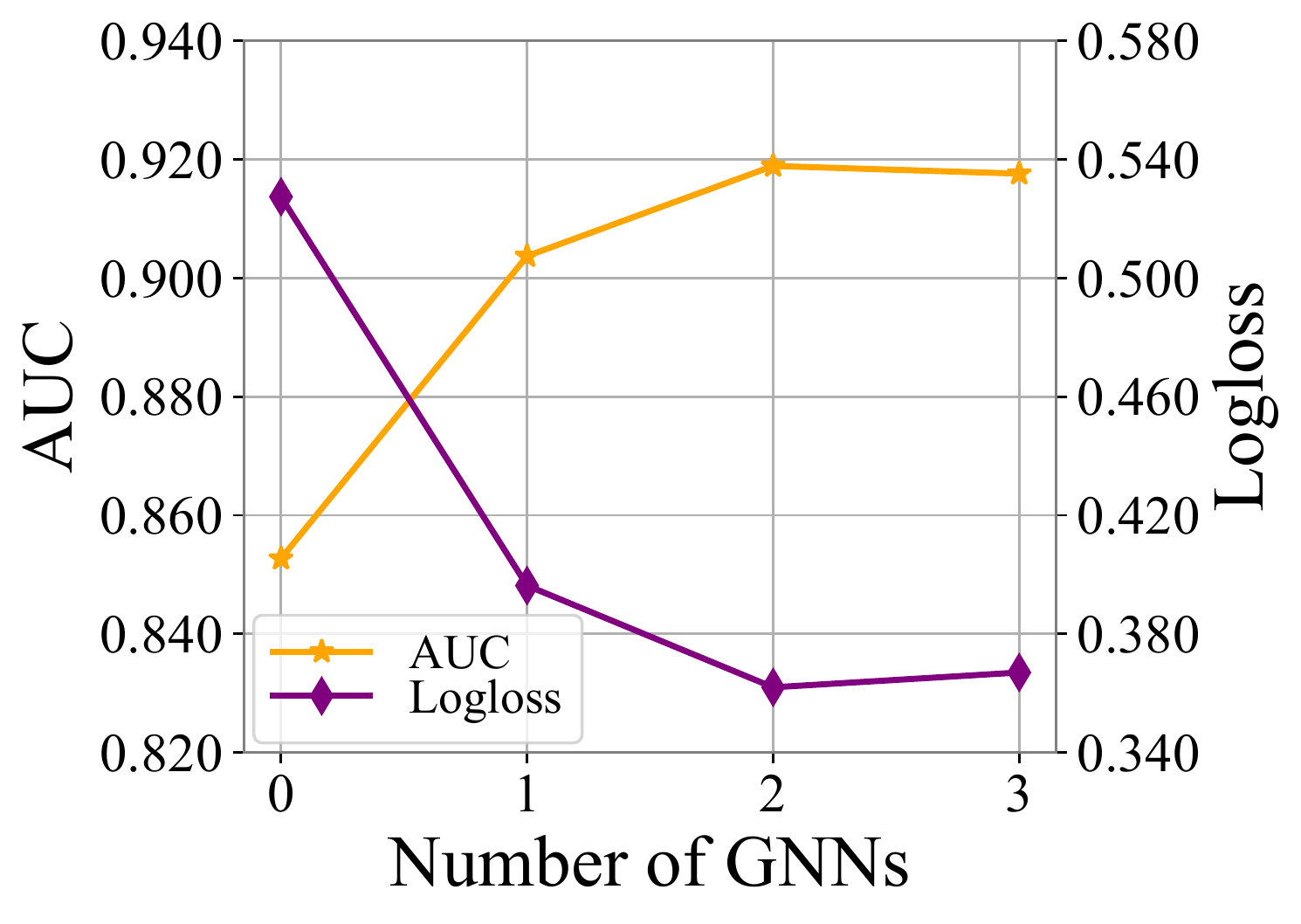}
    \label{fig:paramb}
\end{minipage}
}

\subfigure[Different walk steps on \textbf{Tokyo}]{
\begin{minipage}{0.49\linewidth}
    \includegraphics[width=\linewidth]{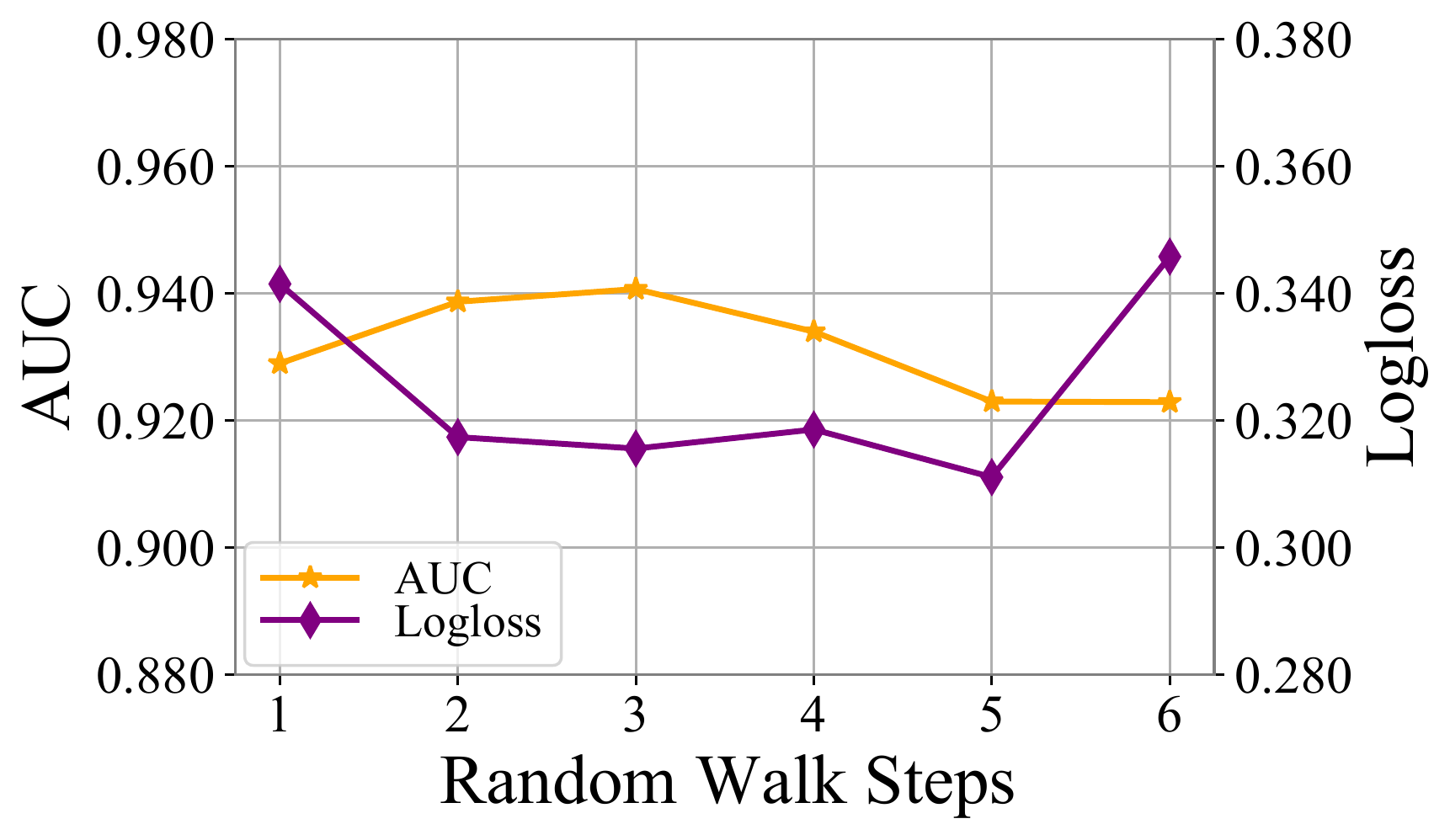}
    \label{fig:paramc}
\end{minipage}
}%
\subfigure[Different walk steps on \textbf{NYC}]{
\begin{minipage}{0.49\linewidth}
    \includegraphics[width=\linewidth]{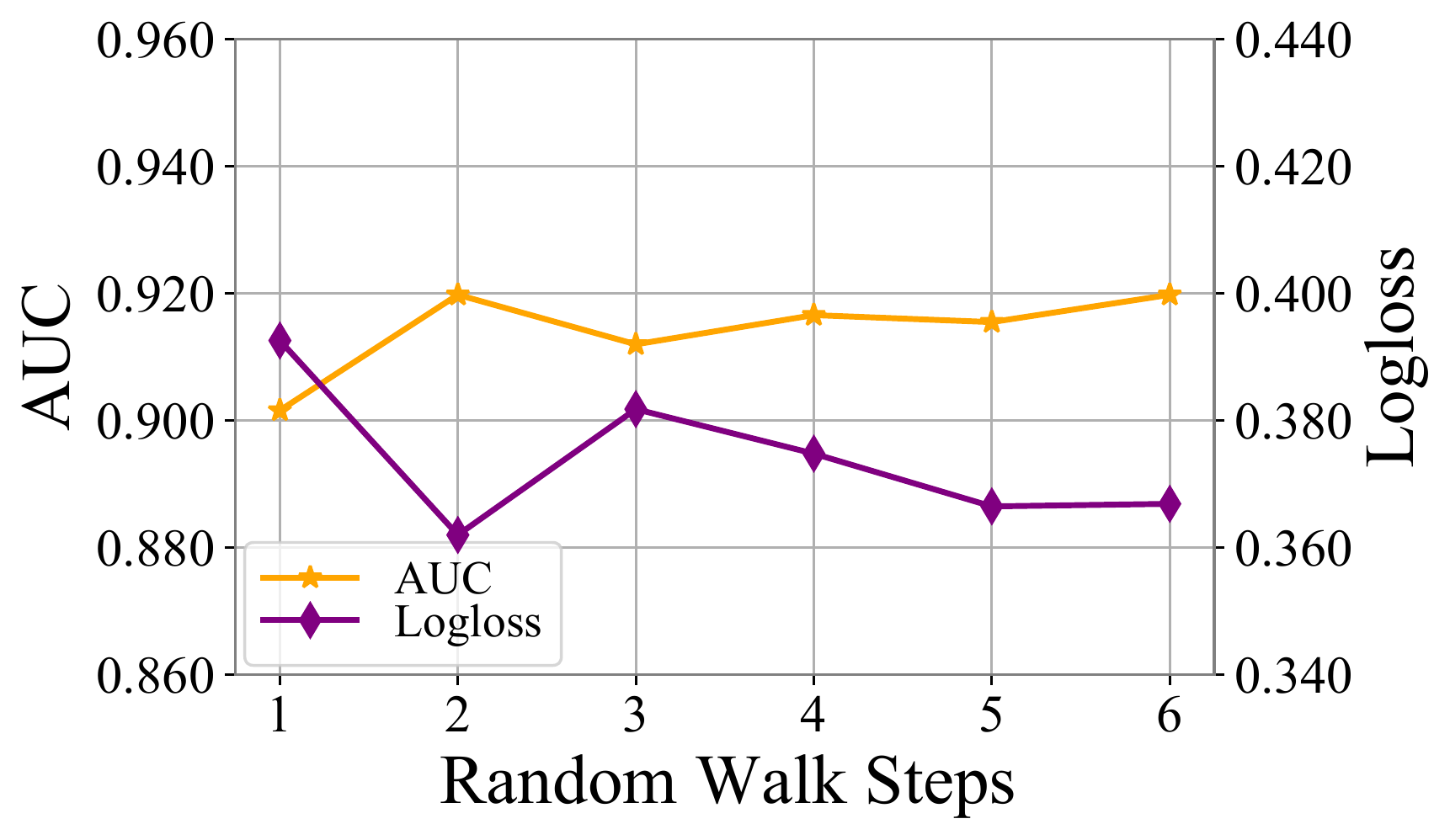}
    \label{fig:paramd}
\end{minipage}
}
\caption{Performance comparison w.r.t. different number of GNN layers and random walk steps.}
\label{fig:param}
\end{figure}

\paragraph{Hyper-parameter Study of Two GNN Modules}
Since we propose to design a message passing graph neural network for the geographical graph and a graph kernel neural network for the sequential graph, it is necessary to study the sensitivity of hyper-parameters within two modules. Specifically, we conduct experiments with different numbers of GNN layers, as well as different steps for random walk graph kernel. From the results in Fig.~\ref{fig:param}, we observe that:
\begin{itemize}[leftmargin=*]
    \item The message passing neural network that encodes topological geographical influences plays an essential role in recommendation. As shown in Fig.~\ref{fig:parama} and Fig.~\ref{fig:paramb}, we can see that the model would degenerate into an attention network when the number of GNN layers is set to $0$ (i.e., without any GNNs), which proves the importance of the message passing neural network for the geographical graph. As the number of GNN layers increases, the performance would first increase to the optimum before saturation, which later decrease to some degree due to the over-smoothing.
    \item The step of random walk would also affect the performance according to Fig.~\ref{fig:paramc} and Fig.~\ref{fig:paramd}. As the random walk step varies from $1$ to $5$, the AUC would first increase and then decrease slowly. The results indicate that a proper random walk step could assist graph kernels in better exploring high-order sequential substructures of users' behaviors. 
\end{itemize}

\paragraph{Influence of Consistency Weight}

\begin{figure}[t]
\centering
\subfigure[Results on \textbf{Tokyo}]{
\begin{minipage}{0.49\linewidth}
    \includegraphics[width=\linewidth]{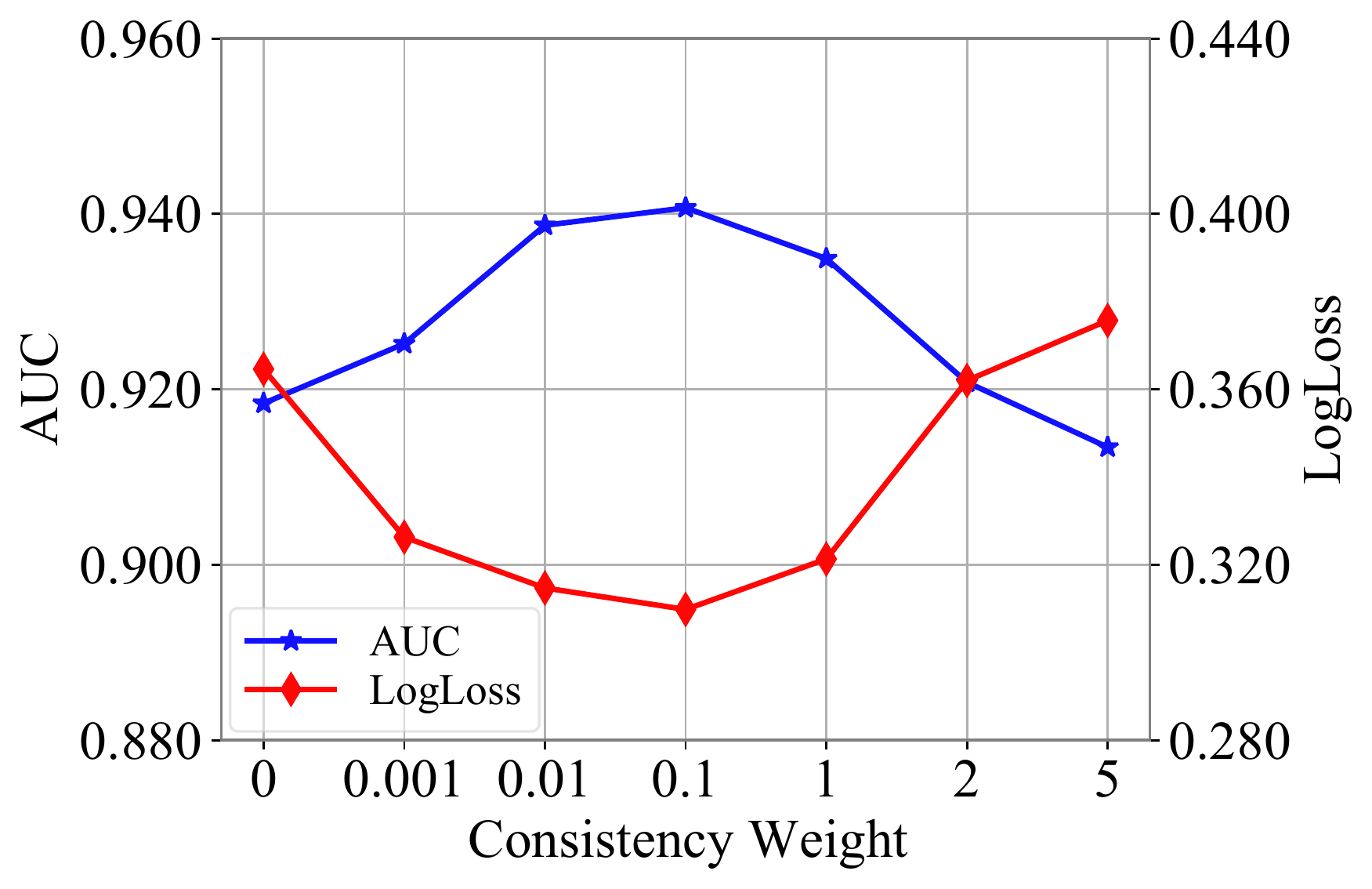}
\end{minipage}
}%
\subfigure[Results on \textbf{NYC}]{
\begin{minipage}{0.49\linewidth}
    \includegraphics[width=\linewidth]{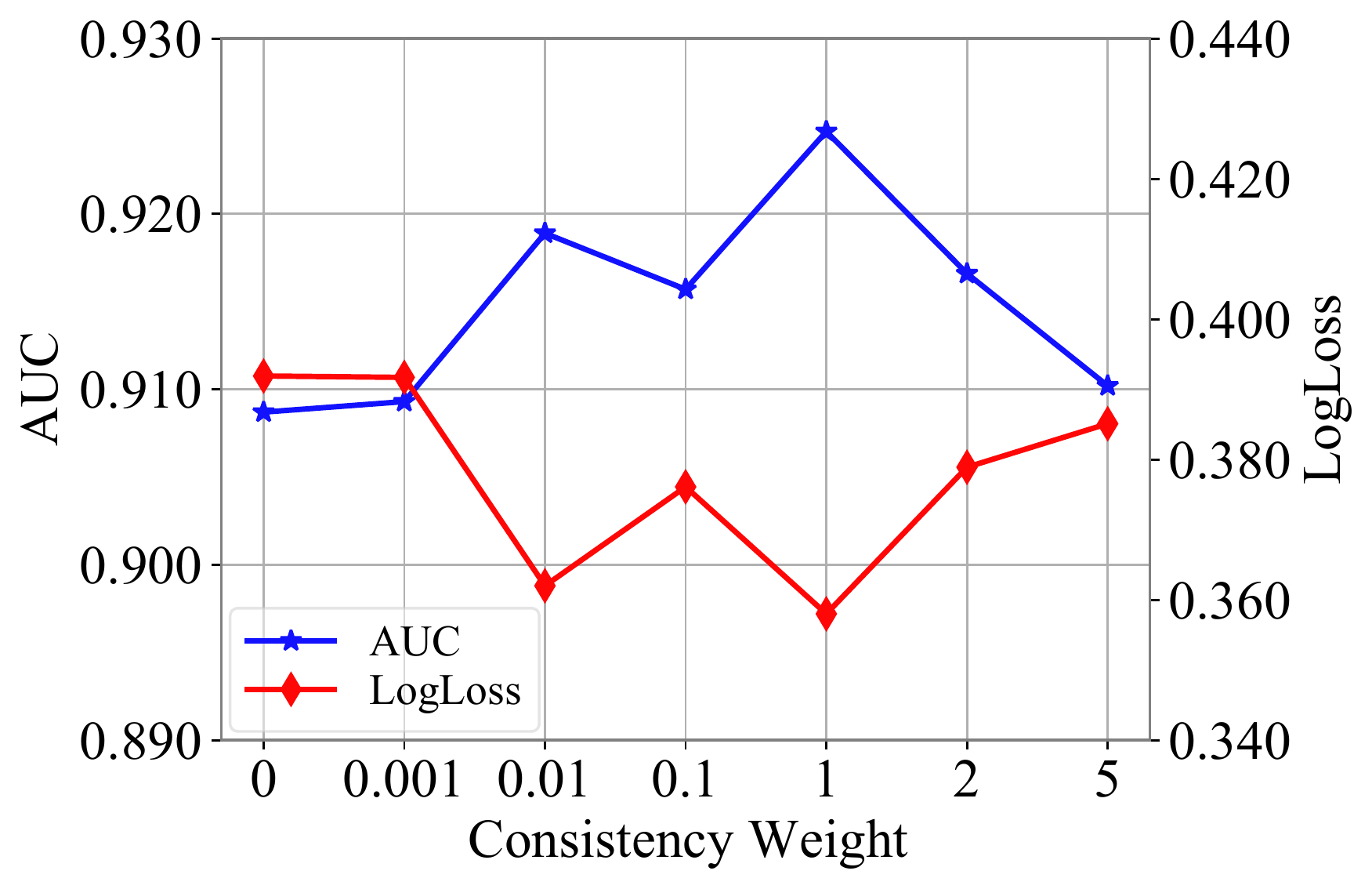}
\end{minipage}
}
\caption{Performance comparison w.r.t. different settings of $\beta$.}
\label{fig:con_weight}
\end{figure}

Since we consider to combine the consistency and recommendation loss in Eq.~\ref{eqn:overall_loss}, we further investigate the influence of the hyper-parameter $\beta$. Specifically, we train the \method{} when $\beta$
varies from $0$ to $5$. From the results shown in Fig.~\ref{fig:con_weight}, it can be seen that:
\begin{itemize}[leftmargin=*]
    \item A larger $\beta$ would encourage stronger consistency between the two GNN modules, which typically leads to better model performance. However, an extreme weight on consistency (greater than $10^{-1}$) will weaken the model capability and deteriorate the recommendation performance.
\end{itemize}

\begin{figure}[t]
\centering
\subfigure[Results on \textbf{Tokyo}]{
\begin{minipage}{0.49\linewidth}
    \includegraphics[width=\linewidth]{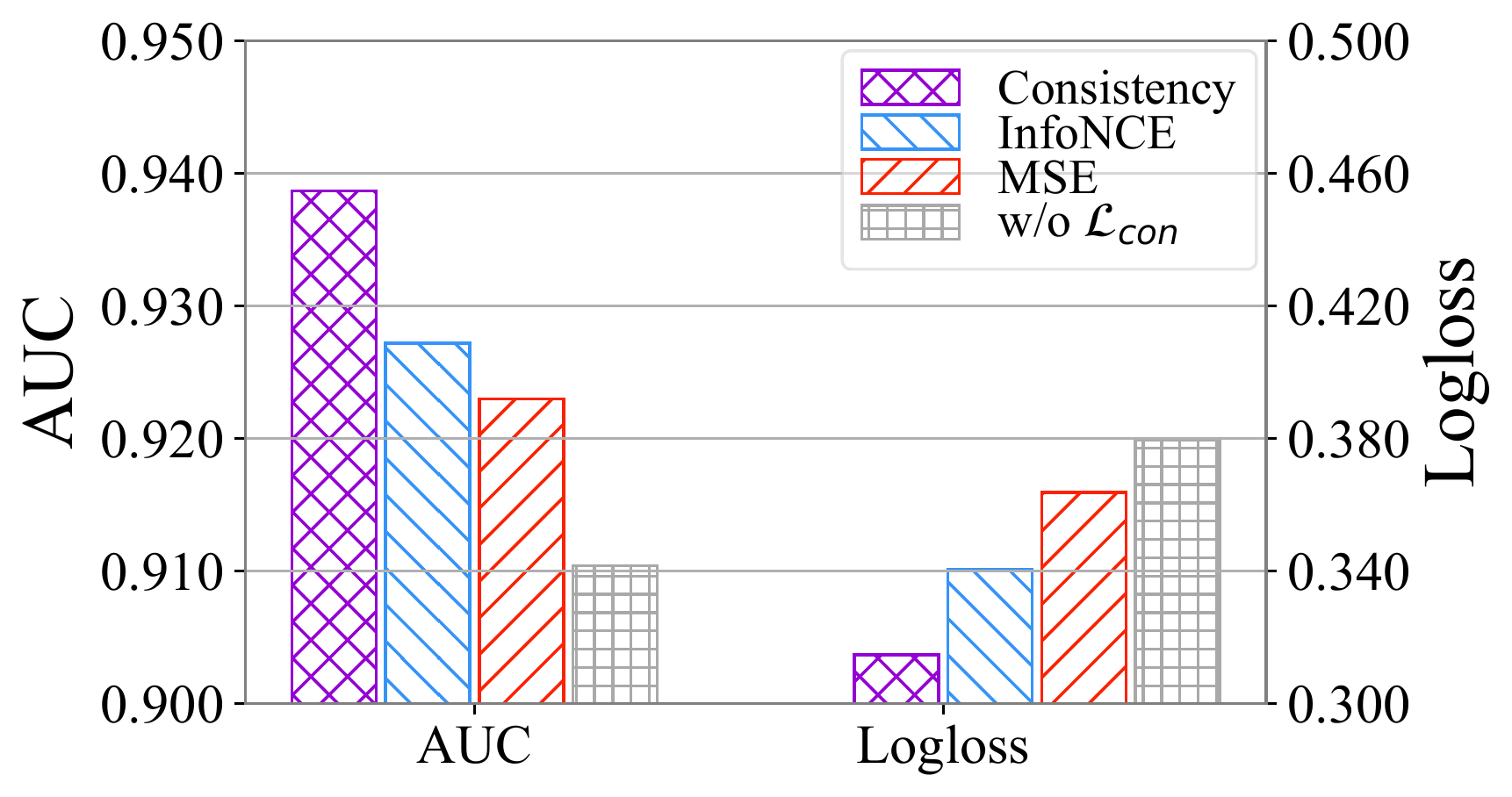}
    \label{fig:cona}
\end{minipage}
}%
\subfigure[Results on \textbf{NYC}]{
\begin{minipage}{0.49\linewidth}
    \includegraphics[width=\linewidth]{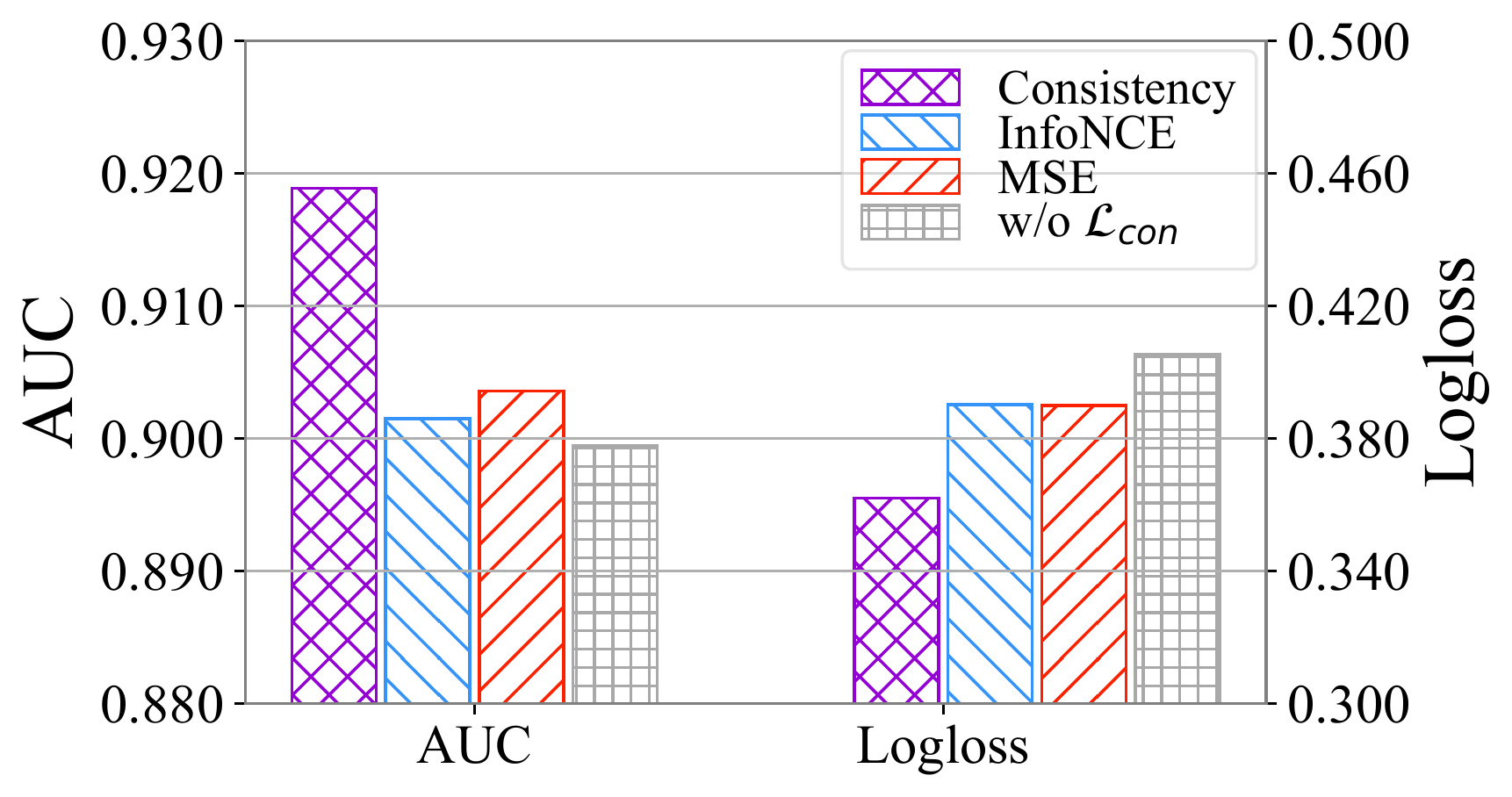}
    \label{fig:conb}
\end{minipage}
}
\caption{Performance comparison w.r.t. different types of $\mathcal{L}_{\text{con}}$.}
\label{fig:con}
\end{figure}

\paragraph{Influence of Consistency Loss}
The consistency loss $\mathcal{L}_{\text{con}}$ is proposed to exchange knowledge between two GNN modules. To further the superiority of the consistency learning framework, we conduct experiments to compare the proposed $\mathcal{L}_{\text{con}}$ with other widely-used contrastive losses. To be specific, we compare the model performance when the $\mathcal{L}_{\text{con}}$ is replaced with InfoNCE loss and mean squared error (MSE) loss respectively. The results are shown in Fig.~\ref{fig:con}. We summarize the results and draw the following conclusions:
\begin{itemize}[leftmargin=*]
    \item The performance of \method{} suffers from the decline significantly when there is no $\mathcal{L}_{\text{con}}$ applied (w/o $\mathcal{L}_{\text{con}}$), compared with the model with different forms of contrastive loss, which suggests that the knowledge intra-communication between two modules is beneficial for recommendation. 
    \item Compared with InfoNCE and MSE, which emphasize on encouraging the similarities between two graph representations in a hard way, our proposed consistency loss achieves the best performance on both datasets. The results indicate that soft alignment between similarity distributions from two views can better encourage consistency learning.
\end{itemize}

\begin{figure}
\centering
\subfigure[Sequential Module]{
\begin{minipage}{0.47\linewidth}
    \includegraphics[width=\linewidth]{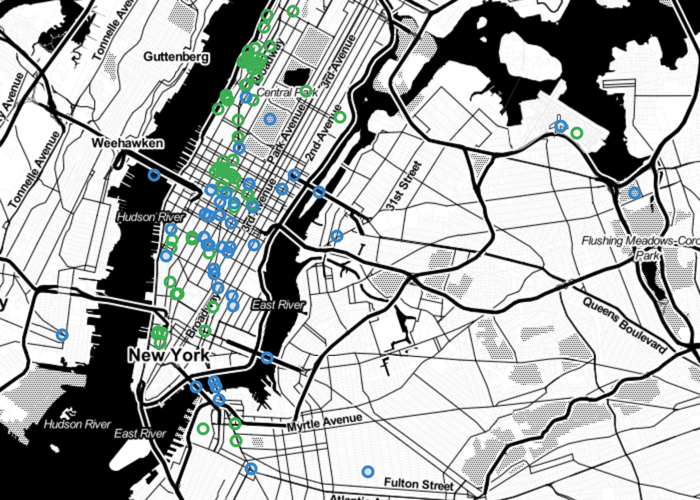}
    \label{fig:mapa}
\end{minipage}
}%
\subfigure[Geographical Module]{
\begin{minipage}{0.47\linewidth}
    \includegraphics[width=\linewidth]{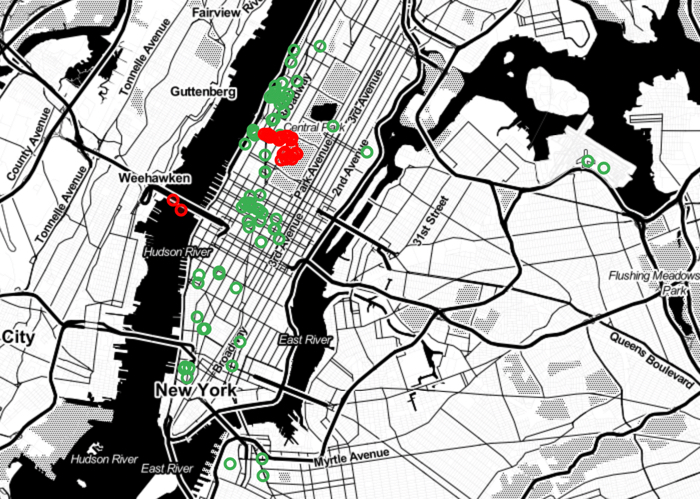}
    \label{fig:mapb}
\end{minipage}
}
\caption{Geographical distribution of the top-recommended next-to-visit POIs. The trajectory of the user 24 is marked with green circles, and the top-scored POIs by two modules are marked with blue and red circles, respectively.}
\label{fig:map}
\end{figure}


\begin{figure}
\centering
\subfigure[]{
\begin{minipage}{0.38\linewidth}
    \includegraphics[width=\linewidth]{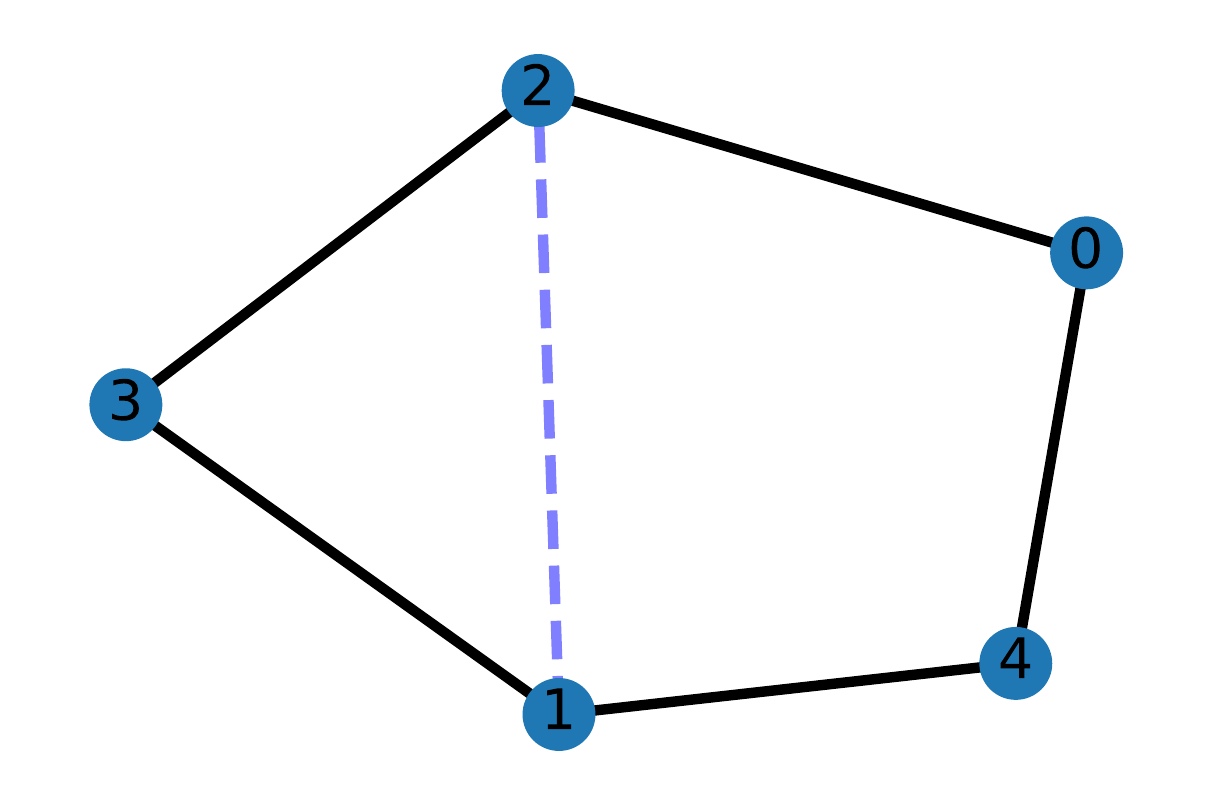}
    \label{fig:case_filter_a}
\end{minipage}
}\ \ \ \ \ \ \ \ \ \ \ \ %
\subfigure[]{
\begin{minipage}{0.38\linewidth}
    \includegraphics[width=\linewidth]{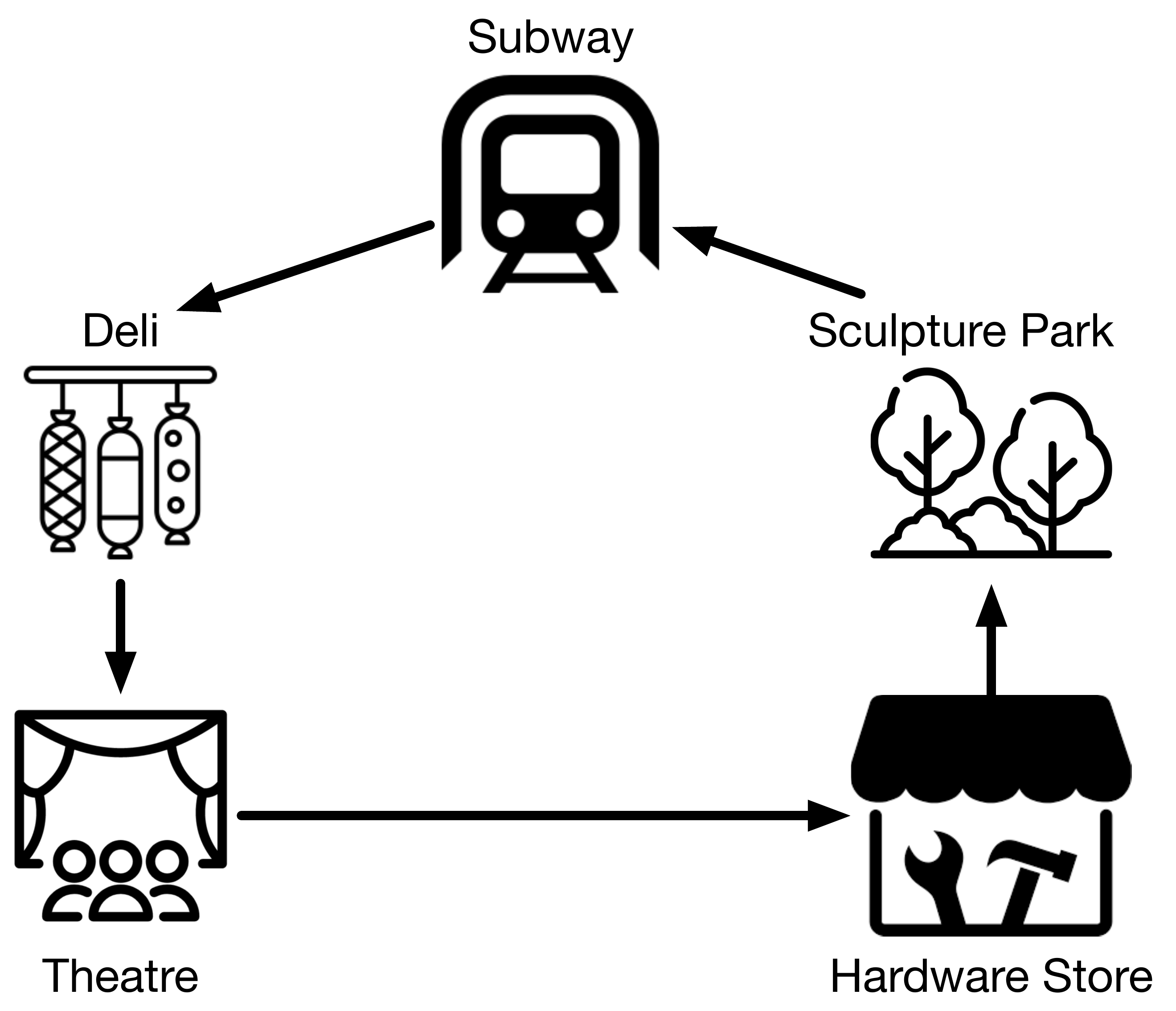}
    \label{fig:case_filter_b}
\end{minipage}
}

\subfigure[]{
\begin{minipage}{0.38\linewidth}
    \includegraphics[width=\linewidth]{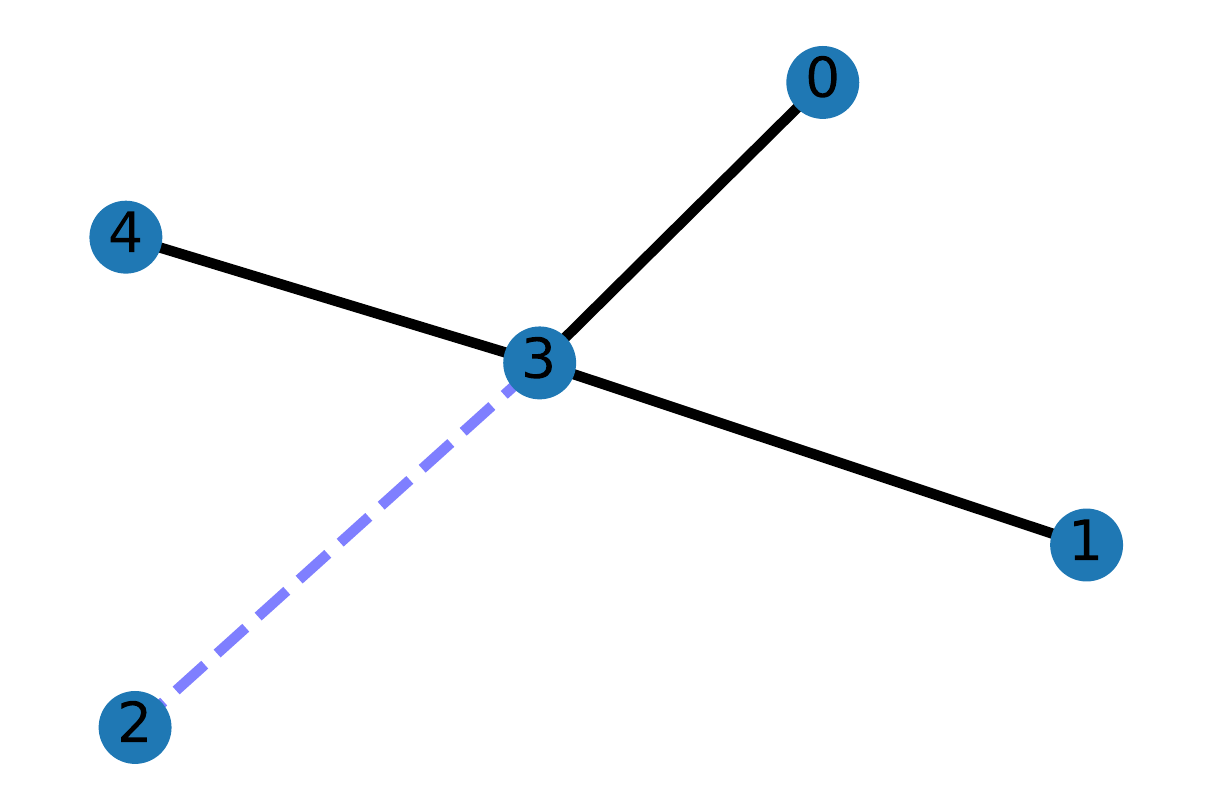}
    \label{fig:case_filter_c}
\end{minipage}
}\ \ \ \ \ \ \ \ \ \ \ \ %
\subfigure[]{
\begin{minipage}{0.38\linewidth}
    \includegraphics[width=\linewidth]{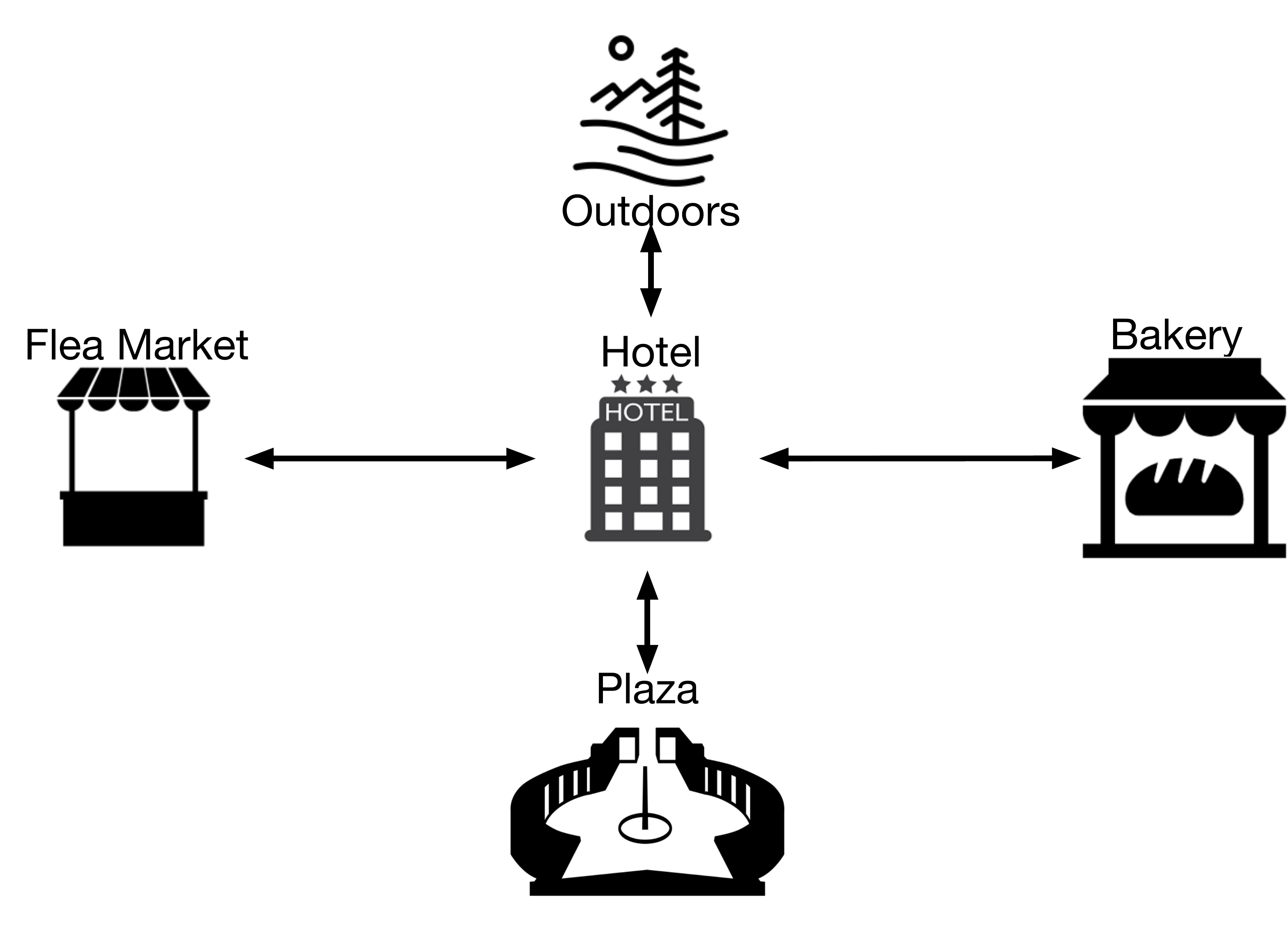}
    \label{fig:case_filter_d}
\end{minipage}
}
\caption{Visualization of graph filters from the sequential module on \textbf{NYC}.}
\label{fig:graph}
\end{figure}

\vspace{-1.5mm}
\subsection{Case Study (RQ3)}

The two modules are built upon different GNN architectures, which empower them to capture respective semantics from two kinds of graphs. We conduct a case study on \textbf{Foursquare NYC} dataset to intuitively illustrate the difference between two modules. We randomly choose the user 24 to visualize his visiting trajectory on the city map of New York as well as the top-50 recommended POIs based on two semantic representations. The results in Fig.~\ref{fig:map} show the difference in the geographical distribution of POIs recommended using two modules. It can be observed that POIs recommended by the sequential module in Fig.~\ref{fig:mapa} tend to disperse around the user's historical visited POIs, while the POIs recommended by the geographical module in Fig.~\ref{fig:mapb} form a clear cluster near the last visited POI of the selected user, which is a market on Amsterdam Avenue. The difference demonstrates that sequential and geographical modules are capable of representing different characteristics of a user's visiting history, thus leveraging information from both modules becomes complementary to make proper recommendations based on both influences.

Furthermore, to show the superiority of capturing high-order sequential substructures via the graph kernel neural network, here we visualize the learned graph filters from user sequences in Fig.~\ref{fig:graph}. Graph filters empower the sequential module with the capability to capture various substructures behind the user behavior sequences. We extract two typical substructure patterns from the user's historical sequence in Fig.~\ref{fig:case_filter_b} and Fig.~\ref{fig:case_filter_d}. The illustrated ``pentagon'' and ``star'' structured substructures can be captured by graph filters shown in Fig.~\ref{fig:case_filter_a} and Fig.~\ref{fig:case_filter_c} respectively. As user preferences may exhibit different patterns, the various graph filters from the sequential module have shown an excellent ability to explore high-order sequential substructures via graph kernels.

\section{Conclusion}
\label{sec::conclusion}

This paper presents a Kernel-Based Graph Neural Network framework (\method{}) for next POI recommendation, which captures geographical and sequential influences from historical behaviors to recommend next-to-visit POIs. To make better use of the knowledge from both sides, \method{} is built upon two modules which encode a pair of complementary graphs to extract geographical and sequential influences respectively. We further introduce a self-supervised loss which encourages the consistency between the two modules via exchanging knowledge to mutually enhance each other. Experiments on two datasets demonstrate the effectiveness of our \method{}. 

\section*{Acknowledgment}
This paper is partially supported by grants from the National Key Research and Development Program of China with Grant No. 2018AAA0101902, and the National Natural Science Foundation of China (NSFC Grant No. 62106008 and No. 62006004). Yifang Qin is partially supported by the National Research Training Program of Innovation for Undergraduates.

\bibliographystyle{IEEEtran}
\bibliography{7-ref}

\end{document}